\documentclass[sigconf,nonacm]{acmart}

\usepackage{mathtools}
\usepackage{subcaption}
\usepackage{tabularx}
\usepackage{array}
\usepackage{multirow}
\usepackage{pbalance}

\newcolumntype{Y}{>{\centering\arraybackslash}X}

\makeatletter
\def\@correspondingauthormark{\g@addto@macro\@currentauthors{%
    \advance\hfuzz by 7pt\relax\textsuperscript{\ding{41}}\relax}}
\makeatother

\newcommand{\V}[1]{\boldsymbol{#1}}          

\begin{document}

\title[Expressive Descriptors of Robot Motion for Nonverbal Human-Robot Communication]{Anatomy of Uncertainty: Expressive Descriptors of Robot Motion for Nonverbal Human-Robot Communication}

\author{Ridhima Bector}
\authornote{These authors contributed equally to this research.}
\affiliation{%
  \institution{Nanyang Technological University}
  \city{}
  \country{Singapore}}
\email{ridhima001@e.ntu.edu.sg}

\author{Souravik Dutta}
\authornotemark[1]
\affiliation{%
  \institution{Nanyang Technological University}
  \city{}
  \country{Singapore}}
\email{souravik001@e.ntu.edu.sg}

\author{Poornima Ramachandran}
\authornote{These authors contributed equally to this research.}
\affiliation{%
  \institution{Nanyang Technological University}
  \city{}
  \country{Singapore}}
  \email{poornima002@e.ntu.edu.sg}

\author{Ree Yan Yeoh}
\authornotemark[2]
\affiliation{%
  \institution{Nanyang Technological University}
  \city{}
  \country{Singapore}}
\email{ryeoh002@e.ntu.edu.sg}

\author{Jui Hien Tan}
\affiliation{%
  \institution{Nanyang Technological University}
  \city{}
  \country{Singapore}}
\email{jtan511@e.ntu.edu.sg}

\author{Domenico Campolo}
\affiliation{%
  \institution{Nanyang Technological University}
  \city{}
  \country{Singapore}}
\email{d.campolo@ntu.edu.sg}

\author{Bernhard Johannes Schmitt}
\affiliation{%
  \institution{Nanyang Technological University}
  \city{}
  \country{Singapore}}
\email{bjschmitt@ntu.edu.sg}

\renewcommand{\shortauthors}{Bector et al.}


\begin{abstract}
Robots working alongside humans must communicate their intended actions together with the uncertainty that arises from incomplete or ambiguous perception. This paper introduces a mathematical framework for expressing perceptual uncertainty through the motion of a robotic manipulator. Drawing on concepts of approach-avoidance and active perception, robot behavior is organized in a Commitment-Vigilance state space whose dimensions are represented through Laban Effort factors, mapping five uncertainty-related states, namely confidence, curiosity, hesitance, fear and inactivity on the uncertainty continuum. A kinematic analysis decomposes goal-directed end-effector motion into a radial task-progress rate and a tangential target-bearing rate, which realize the two dimensions. From this decomposition, five motion primitives, namely approach, pause, retreat, probe and twitch, are derived and parameterized using eleven kinematic descriptors covering approach and retreat characteristics, pause behavior, gaze angles, end-effector tilt and shiver amplitude. A video-based human-subject study evaluated the recognition of uncertainty-expressive trajectories and the influence of individual descriptors on perceived intensity. For every trajectory the intended behavioral state was the modal response and was selected significantly more often than chance. In single-descriptor comparisons, participants significantly preferred one variant as the more intense expression of the intended state. The results provide a perceptual basis for encoding robot uncertainty in motion and for generating such trajectories autonomously from parametric movement representations. Expressive robot motion videos and questionnaire used in the user study are available at \url{https://anonymous.4open.science/r/aou/}.
\end{abstract}


\maketitle


\section{Introduction}
\label{sec:Introduction}

Robots operating in human environments rarely possess complete or error-free knowledge of the world. Sensor noise, occlusion, ambiguous object observations, imperfect models, and changes in the task environment produce uncertainty about the current state and the consequences of future actions. Classical decision-making formulations represent this condition through a belief over possible world states \cite{Kaelbling1998}, while active-perception methods allow a robot to select movements that improve its observations and reduce uncertainty \cite{Bajcsy1988}. In human-robot collaboration, uncertainty is not solely an internal estimation problem. A human collaborator must also be able to infer whether the robot has confidently resolved the situation, is gathering additional information, or is unable to determine an appropriate action. A robot that executes every action with an identical motion profile may consequently appear more certain and capable than its perceptual state warrants. However, collaborative-application requirements, such as bounded speed, separation distance and contact force (incorporated into ISO~10218-2:2025 \cite{ISO10218-2}), do not specify how a robot should communicate its intent or confidence to the humans working beside it \cite{Villani2018}.

Robot motion provides an immediately available channel through which such internal states can be communicated. Nonverbal robot behaviors have been shown to improve the understandability, efficiency, and robustness of collaborative interaction \cite{Breazeal2005}. Animation principles such as anticipation and reaction can make robot actions easier to interpret \cite{Takayama2011}, while legible-motion formulations explicitly optimize trajectories to reveal an intended goal \cite{Dragan2013}. More broadly, motion-design research has argued that timing, rhythm, posture, and transitions should be treated as communicative properties rather than as incidental consequences of control \cite{Hoffman2014}. On the industrial front, Apple's ELEGNT lamp-like robot has combined functional and expressive movement primitives to convey intention, attention and emotion through a non-anthropomorphic body \cite{Hu2025}. These findings are particularly relevant to robotic manipulators, which frequently lack anthropomorphic faces, speech interfaces, or dedicated displays but continuously produce observable motion during task execution. Their functional trajectories can therefore act simultaneously as task actions and as nonverbal signals about attention, intention, and internal state.

A remaining central challenge is determining how motion should change when the robot is perceptually uncertain. Laban Movement Analysis (LMA) is a comprehensive framework with systematic vocabulary for describing and interpreting human movement \cite{Laban1947,Bartenieff1980}. LMA-derived features have conveyed affect through quadrotor flight paths \cite{Sharma2013}, a mobile robot's attitude toward its goal through path shape \cite{Knight2014,Knight2016} and emotion through manipulator motion \cite{LaViola2022}. Independent work on expressive motion timing has further demonstrated that changes in speed and pauses can alter perceived robot confidence even when the geometric path remains fixed \cite{Zhou2017}. Nevertheless, most existing methods map selected motion qualities to discrete emotions, personality traits, or intended goals. They do not provide a general mathematical account of how perceptual uncertainty should be decomposed into behavioral processes or identify a sufficient set of measurable motion parameters through which different uncertainty-related states and their intensities can be expressed.

In this work, we represent perceptual uncertainty along two dimensions. The first dimension is drawn from approach-avoidance motivation \cite{Elliot2006}, i.e. how far the robot commits its motion toward or away from the goal. The second dimension is based on active and interactive perception \cite{Bajcsy1988}, i.e. how much of its motion it allocates to monitoring and information gathering. Our principal contributions are as follows.

\begin{enumerate}
    \item \textbf{An anatomical behavioral model of perceptual uncertainty}, in which the four Laban Effort factors are mathematically projected onto two orthogonal dimensions. The resulting state space organizes confidence, curiosity, hesitance, fear, and inactivity according to the robot's task-directed engagement and uncertainty-management behavior.
    \item \textbf{A set of motion primitives and computable expressive descriptors} that translate the behavioral model into manipulator trajectories. Goal-directed end-effector motion is decomposed into a task-progress rate and a target-bearing rate, from which five motion primitives, namely approach, pause, retreat, probe, and twitch, are derived and parameterized using eleven kinematic descriptors.
    \item \textbf{A human-perception validation of the proposed representation} examining whether descriptor-parameterized trajectories communicate distinct uncertainty-related behavioral states and whether systematic changes in individual descriptors modulate their perceived expressive intensity.
\end{enumerate}


\section{Related Work}
\label{sec:RelatedWork}

\subsection{Expressive Robot Motion}
\label{subsec:ExpressiveRobotMotion}

Expressive robot motion encompasses behaviors designed not only to achieve a physical objective but also to make the robot's intentions, affect, capabilities, or interaction state understandable to an observer. Early work showed that implicit nonverbal cues can improve coordination and recovery from miscommunication in human-robot teamwork \cite{Breazeal2005}. Takayama et al. applied the animation principles of anticipation and reaction to make robot actions more readable \cite{Takayama2011}, while Dragan et al. distinguished predictable motion, which conforms to an observer's expectation, from legible motion, which enables rapid inference of the robot's goal \cite{Dragan2013}. Hoffman and Ju subsequently framed movement as a fundamental robot-design material, emphasizing that temporal phrasing and transitions influence how humans interpret even mechanically simple platforms \cite{Hoffman2014}. Recently, Hielscher et al. implemented animation principles utilizing Dynamic Movement Primitive (DMP) parameters to generate expressive modulation of robot motion \cite{Hielscher2025}. In industry, animation principles also drive artist-directed motion on entertainment robots \cite{Grandia2024}. Complementary studies have investigated external and embodied methods for communicating intent in shared workspaces, including trajectory modifications, gaze, lights, projections, and augmented-reality cues \cite{Pascher2023,Lemasurier2021}.

Within this broader area, LMA has provided a structured representation for transferring principles of human expressive movement to nonhuman embodiments. Burton et al. proposed LMA as a compact abstraction for affective-motion recognition and generation \cite{Burton2015}. Sharma et al. encoded Laban Effort qualities into quadrotor flight paths and showed that observers could perceive different affective characteristics from locomotion alone \cite{Sharma2013}. Knight and Simmons converted Effort factors into path and velocity features for differential-drive robots, demonstrating that low-degree-of-freedom platforms can express attitudes such as confidence, urgency, and reluctance \cite{Knight2014}. Expressive timing has also been studied independently of path geometry where Zhou et al. found that pauses and temporal profiles influenced perceived confidence, naturalness, and even assumptions about the object being manipulated \cite{Zhou2017}. More recently, La Viola et al. applied Laban-inspired timing, path, flow, and joint configurations to a robotic arm and obtained recognizable emotional interpretations from observers \cite{LaViola2022}.

These studies establish that humans infer meaning from robot velocity, acceleration, path shape, posture, pauses, and movement rhythm. Nevertheless, the represented meanings are commonly fixed emotional labels, personality styles, task goals, or general competence impressions. LMA factors are frequently implemented as qualitative presets, and the selected features are often specific to a particular robot morphology or experimental scenario. Consequently, the literature does not yet provide a unified decomposition that explains \textit{why} perceptual uncertainty should produce particular motion changes or \textit{how} several changes can be composed to represent qualitatively different uncertainty-related states.

\subsection{Robot Uncertainty and Its Communication}
\label{subsec:RobotUncertainty}

Existing approaches to robot-uncertainty communication span verbal, visual, and behavioral modalities. Bartneck and Moltchanova investigated linguistic probability expressions such as ``likely'' and ``almost certainly,'' finding systematic but imperfect correspondence between verbal phrases and numerical probabilities \cite{Bartneck2020}. Visual interfaces can present confidence scores, distributions, or uncertainty indicators explicitly; Sch\"{o}mbs et al. studied uncertainty visualization in robot-assisted decision-making and demonstrated that both the representation and embodiment of uncertainty influence user judgments \cite{Schoembs2024}. Such explicit methods can communicate quantitative information, but they require users to attend to an additional display or interpret statistical terminology, and they may be less suitable when attention is divided across a physical collaborative task.

Behavioral signals provide an alternative in which uncertainty is conveyed through the robot's actions. Van den Brule et al. showed that nonverbal warning signals preceding poor performance can improve interaction and support more appropriate expectations of the robot \cite{VandenBrule2016}. Kwon et al. formulated the expression of robot incapability as a trajectory-optimization problem, generating task-specific motions that communicate both the intended action and the physical reason that it cannot be completed \cite{Kwon2018}. These approaches demonstrate the value of exposing limitations before or during failure, but incapability and expected performance are not identical to perceptual uncertainty: a robot may remain physically capable while being unsure about an object, observation, or appropriate course of action.

Hesitation has been the uncertainty-related motion behavior studied most directly. Moon et al. derived an acceleration-based hesitation profile from human reaching motions and implemented it on a robot arm during shared-resource conflicts \cite{Moon2013}. Later, they proposed the Negotiative Hesitation Generator, which reproduced characteristic human-inspired stopping and yielding trajectories and showed that such behavior can facilitate nonverbal conflict negotiation \cite{Moon2021}. Expressive timing research similarly found that pauses can make a manipulator appear uncertain or as though it is reconsidering its action \cite{Zhou2017}. Recent Laban-based manipulator work has introduced hesitant trajectories using sustained timing, retreating Shape qualities, and direct or curved movement paths, with observers associating these motions with reduced competence and task-success likelihood \cite{Raghu2025}.

The uncertainty literature therefore confirms that hesitation, delays, warning cues, retreats, and displays can affect human interpretation. However, most methods encode uncertainty as a binary condition, a single probability, an impending failure, or one context-specific hesitation gesture. They seldom distinguish confidence, information-seeking curiosity, approach-withdrawal conflict, defensive fear, and complete disengagement within a common model. Furthermore, relatively little work identifies which independent kinematic variables control the perceived intensity of these states. The methodology developed in this manuscript addresses these limitations by representing perceptual uncertainty as a structured continuum and by specifying a reusable set of trajectory-level descriptors through which uncertainty-related behaviors can be generated, measured, and systematically varied.


\section{Methodology}
\label{sec:Methodology}

We start from the premise that uncertainty in a robot's perception of the
world can be communicated not only through task delay or trajectory error but
through systematic changes in motion quality. Section~\ref{subsec:CVModel} develops a general behavioral model of uncertainty from approach-avoidance and appraisal theories, and makes it
computable by expressing its two dimensions through Laban Effort factors. The resulting state space
organizes uncertainty-related behavior into behavioral classes and canonical
states. Section~\ref{subsec:KinematicBasis} analyzes the kinematics of
goal-directed manipulator motion and shows that the end-effector velocity
decomposes into a radial task-progress rate and a tangential target-bearing
rate, which realize the dimensions of the behavioral state-space as measurable properties of a
robot trajectory. Section~\ref{subsec:ExpressiveDescriptors} uses this decomposition
to derive the motion primitives constituting the canonical states
and the expressive descriptors that parameterize them.

\subsection{Behavioral Modeling of Robot Uncertainty}
\label{subsec:CVModel}

Appraisal theory evaluates the certainty of a
situation separately from the coping potential it affords \cite{Scherer2001}.
An agent that acts under perceptual uncertainty manages two processes at
once, viz., its action tendency toward or away from the goal, and the share of its
behavior that it allocates to monitoring and to reducing the uncertainty.
Motivation research treats approach and avoidance as the basic directions of
behavior \cite{Elliot2006}. In robotics, control policies under partial observability 
contain actions whose only return is reduced uncertainty \cite{Kaelbling1998}, 
and active and interactive perception treat motion as a means of acquiring information \cite{Bajcsy1988,Bajcsy2018,Bohg2017}. 
A robot must therefore decide at every instant where to move relative to the goal and
how much motion to spend finding out.

We formalize these two processes as the orthogonal dimensions of
\textbf{Commitment}, $C \in [-1, 1]$, and \textbf{Vigilance},
$V \in [-1, 1]$. Commitment measures whether the robot invests motion toward
task completion or withdraws from it. Positive commitment appears as
advancing, direct, target-oriented motion; zero commitment appears as
stopping; negative commitment appears as retracting or yielding. Vigilance measures how much of the robot's motion is spent on uncertainty management rather than on direct execution.
Low vigilance corresponds to narrow, stable target fixation and uninterrupted
exploitation; high vigilance corresponds to active information seeking or
defensive monitoring, expressed as gaze sweeps, orientation adjustments or
pauses for checking.

\subsubsection{Laban Effort factors as a computable basis}

To make the two dimensions computable, and to relate them to the quality of
movement, we draw on the Laban Effort system. While it describes movement by four
motion factors, each associated in the LMA literature with a distinct inner
attitude \cite{Laban1947,Bartenieff1980}, we interpret each in terms of
robot uncertainty processing. The \textit{Weight} ($W$) factor
(Strong~$\leftrightarrow$~Light) encodes intentional engagement, i.e., the
resolution of force toward action. The \textit{Time} ($T$) factor
(Sudden~$\leftrightarrow$~Sustained) captures decision urgency, the temporal
pressure under which the agent acts. The \textit{Flow} ($F$) factor
(Bound~$\leftrightarrow$~Free) reflects regulation, the degree to which motion
is controlled rather than released. The \textit{Space} ($S$) factor
(Direct~$\leftrightarrow$~Indirect) governs attentional focus, the agent's
mode of environmental scanning.

We represent the Effort quality of a movement by the normalized
four-dimensional vector $E = (W, T, F, S) \in [-1, +1]^4$ and adopt the sign
convention of the EMOTE model \cite{Chi2000}, where each factor spans from its
\textit{indulging} pole at $-1$ to its \textit{fighting} pole at $+1$. Strong
Weight, Sudden Time, Bound Flow and Direct Space take the value
$+1$, and Light Weight, Sustained Time, Free Flow and Indirect Space the value
$-1$.

Commitment and Vigilance are each derived from a pairing of Effort factors
that corresponds to one of Laban's incomplete-effort States
\cite{Bartenieff1980}. Commitment is anchored in the \textit{Near} (rhythm)
State, which combines Weight and Time, the factors that concern the exertion
of force toward action and its timing. Vigilance is anchored in the
\textit{Remote} State, which combines Space and Flow, the factors that concern
attention and the control of movement:

\begin{align}
  C &= c_1W + c_2T;   & W, T \in [-1, +1] \label{eq:CDimension} \\
  V &= v_1F - v_2S;  & F, S \in [-1, +1] \label{eq:VDimension}
\end{align}

Here $c_1, c_2, v_1, v_2 \in \mathbb{R}^+$ are weighting coefficients with
$c_1 + c_2 = v_1 + v_2 = 1$, which keeps $C$ and $V$ within $[-1, 1]$. Strong Weight
($W = +1$) and Sudden Time ($T = +1$) yield maximal Commitment, while Light
Weight ($W = -1$) and Sustained Time ($T = -1$) yield minimal Commitment.
Bound Flow ($F = +1$) and Indirect Space ($S = -1$) yield maximal Vigilance,
while Free Flow ($F = -1$) and Direct Space ($S = +1$) yield minimal
Vigilance.

\subsubsection{Behavioral classes and canonical states}

\begin{figure}[tbp]
  \centering
  \includegraphics[width=\columnwidth]{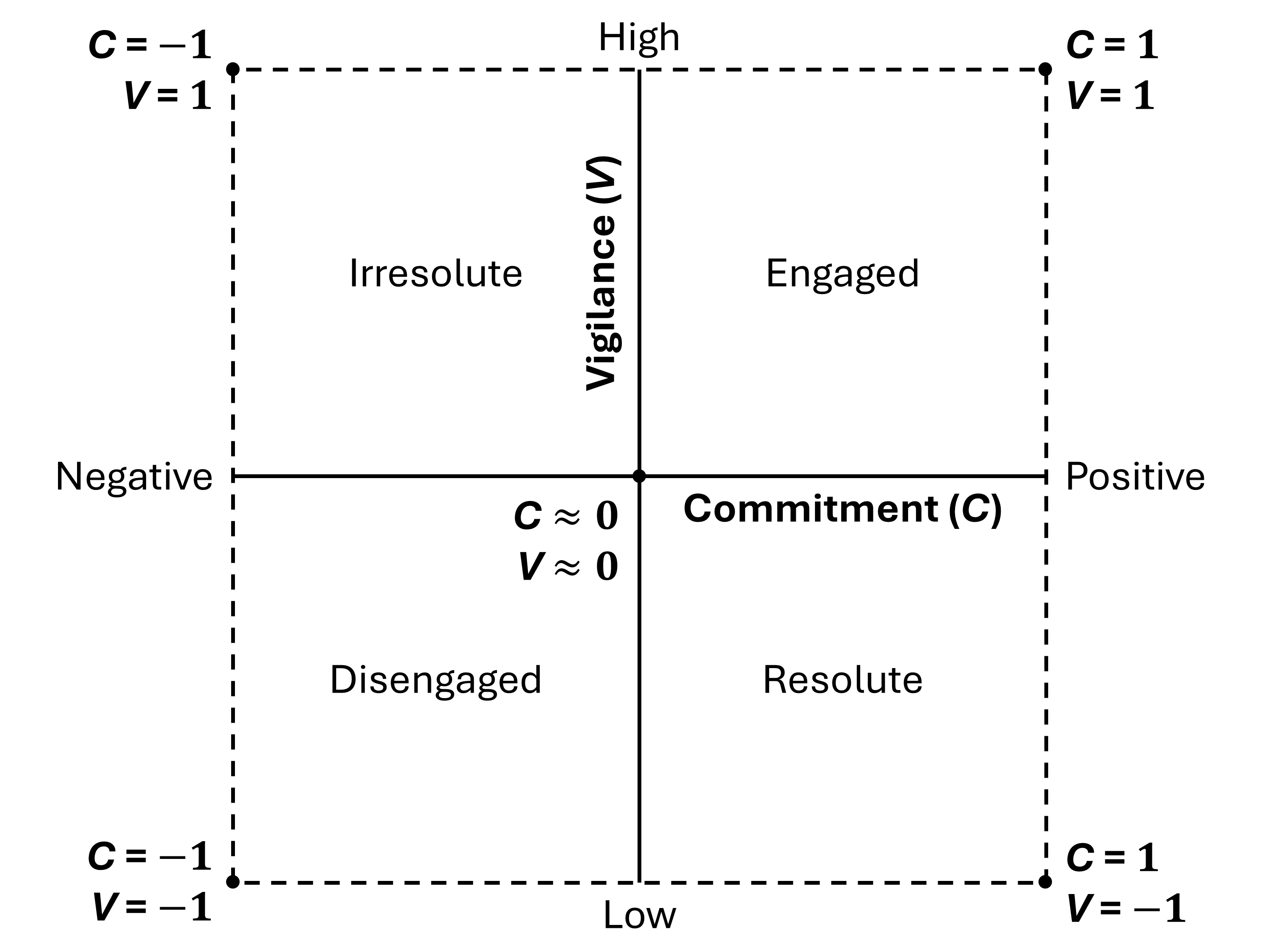}
  \caption{Commitment-Vigilance state space for robot uncertainty. Quadrants depict the four          behavior classes with the corners and the origin marking the five canonical states, namely
    Pure Confidence ($C = 1$, $V = -1$), Pure Curiosity ($C = 1$, $V = 1$), Pure Hesitance
    ($C \approx 0$, $V \approx 0$), Pure Fear ($C = -1$, $V = 1$), and Inactivity
    ($C = -1$, $V = -1$).}
  \Description{A two-dimensional plane with Commitment on the horizontal axis,
    ranging from negative on the left to positive on the right, and Vigilance
    on the vertical axis, ranging from low at the bottom to high at the top.
    The four quadrants are labeled Irresolute (top left), Engaged (top
    right), Disengaged (bottom left) and Resolute (bottom right). The corners
    are annotated with the corresponding extreme values of C and V, and the
    origin is annotated with C approximately zero and V approximately zero.}
  \label{fig:CVModel}
\end{figure}

Fig.~\ref{fig:CVModel} shows the resulting state space. It partitions into
four behavioral classes with respect to a goal position or object.
Positive commitment with low vigilance ($C > 0$, $V < 0$) describes a
\emph{resolute} agent that acts decisively without defensive monitoring.
Positive commitment with moderate or high vigilance ($C > 0$, $V \geq 0$) describes
an \emph{engaged} agent that approaches while scanning. Zero or
negative commitment with moderate or high vigilance ($C \leq 0$, $V \geq 0$) describes an
\emph{irresolute} agent that withdraws or holds back under uncertainty.
Zero or negative commitment with low vigilance ($C \leq 0$, $V < 0$)
describes a \emph{disengaged} agent that is neither acting nor monitoring.

Within these classes, we define five canonical states of increasing perceptual uncertainty. Each is
characterized by two appraisals \cite{Scherer2001} -- certainty, i.e., how
resolved the assessment of the environment is, and coping potential, i.e.,
how capable the agent judges itself of engaging. Pure \textbf{Confidence} ($C = 1$, $V = -1$) combines maximal certainty with high coping potential -- the
robot has resolved its assessment and commits fully. Pure \textbf{Curiosity} ($C = 1$, $V = 1$)
combines moderate uncertainty with maintained coping potential -- the
assessment is unresolved, but the situation appears rewarding enough to
warrant probe. Pure \textbf{Hesitance} ($C \approx 0$, $V \approx 0$) combines high uncertainty with low coping potential -- the robot detects a possible threat, cannot decide between
approach and withdrawal, and oscillates between them. Pure \textbf{Fear} ($C = -1$, $V = 1$) combines high uncertainty with no coping potential -- the environment is judged
threatening and the capacity to engage insufficient. \textbf{Inactivity} ($C = -1$, $V = -1$)
combines maximal uncertainty with no coping potential -- the robot disengages
from the task.

The extreme coordinates of the $C$-$V$ state space represent the dominant quality of each behavior over a single motion segment. Across a complete task, however, these behavioral states do not occur in isolation; they combine according to task progression and the robot's management of uncertainty, and they frequently overlap to produce mixed behaviors. Consequently, the uncertainty-expressive behavioral states of confidence, curiosity, hesitance, and fear observed in real-world robot tasks rarely coincide with the pure cases at the extremes and instead occupy intermediate positions within the quadrants of the $C$-$V$ state space, admitting various Effort signature combinations.

\subsection{Kinematic Decomposition of Uncertainty-Expressive Robot Motion}
\label{subsec:KinematicBasis}

The $C$-$V$ model specifies what uncertainty-expressive behavior must convey
but not which components of a robotic manipulator's motion convey it. We therefore
analyze the kinematics of a manipulator in a goal-directed task and show that
its end-effector velocity has a unique orthogonal decomposition relative to the
goal, whose two components are independently commandable and correspond to
Commitment and Vigilance. Uncertainty-expressive motion can then be treated
anatomically, as a composition of kinematically distinct parts.

\begin{figure*}[t]
  \centering
  \begin{subfigure}[b]{0.3172\textwidth}
    \centering
    \includegraphics[width=\linewidth]{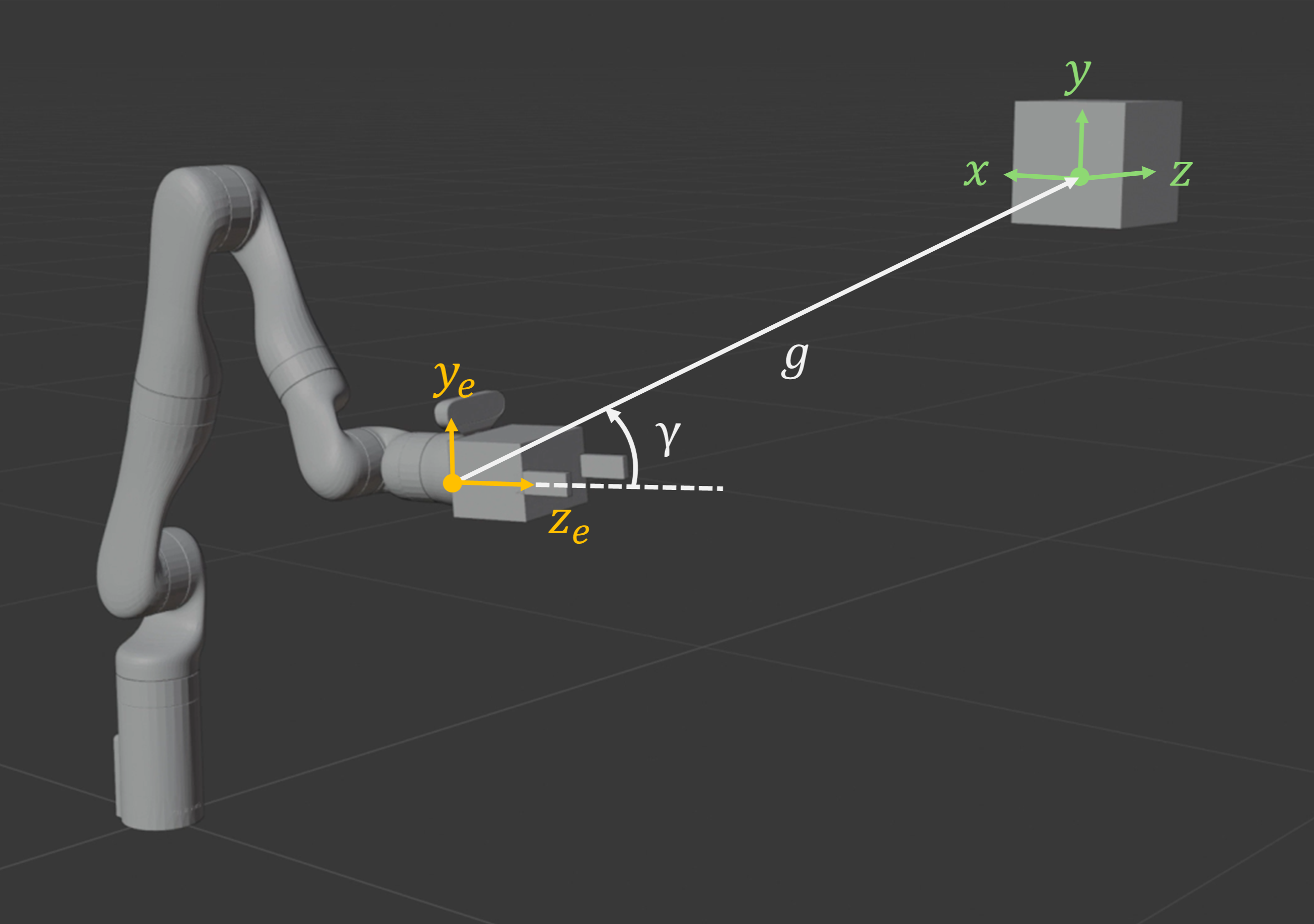}
    \caption{Arbitrary robot configuration}
    \label{fig:GeometryArbitrary}
  \end{subfigure}\hfill
  \begin{subfigure}[b]{0.3172\textwidth}
    \centering
    \includegraphics[width=\linewidth]{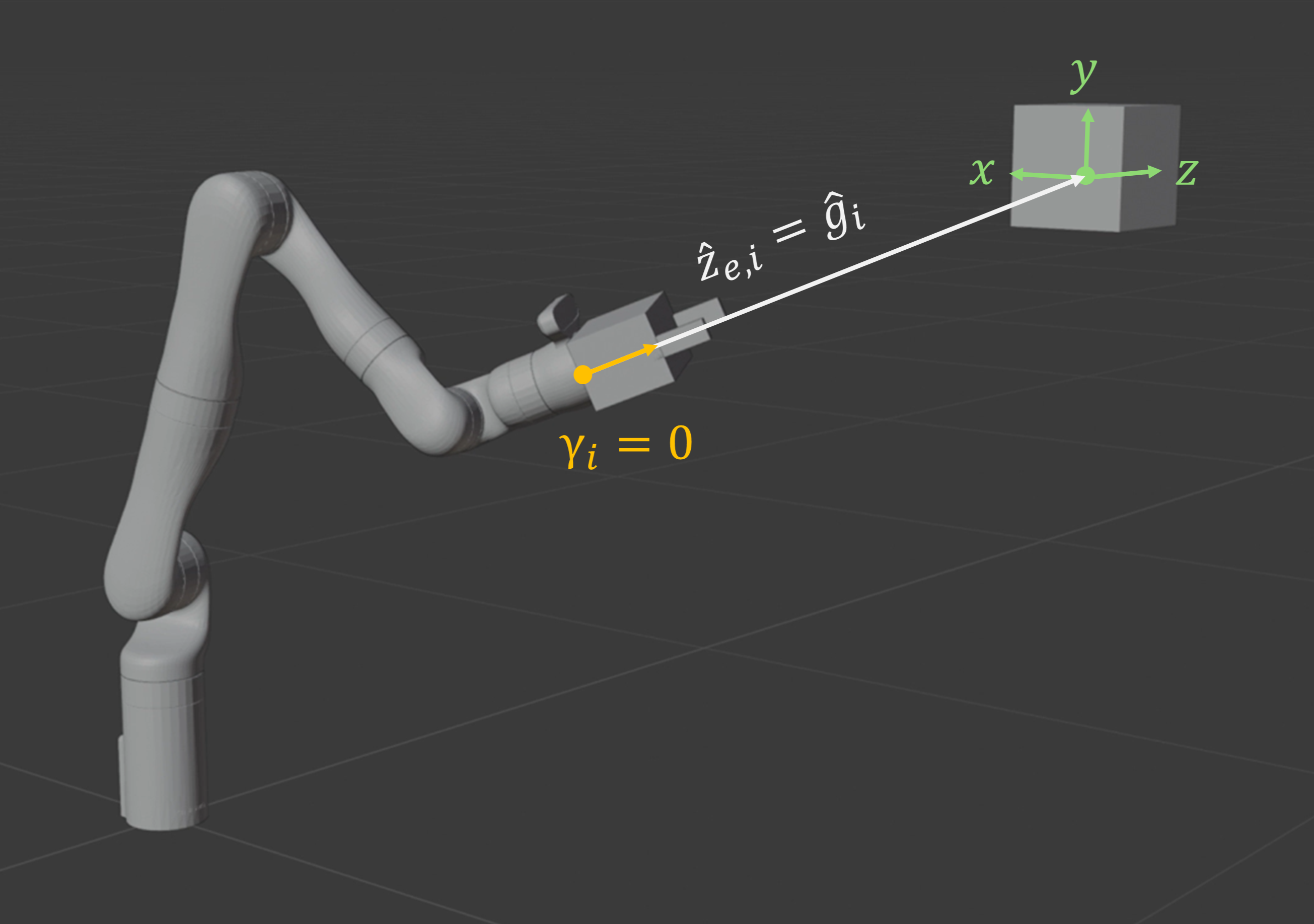}
    \caption{Operational robot configuration}
    \label{fig:GeometryOperational}
  \end{subfigure}\hfill
  \begin{subfigure}[b]{0.3454\textwidth}
    \centering
    \includegraphics[width=\linewidth]{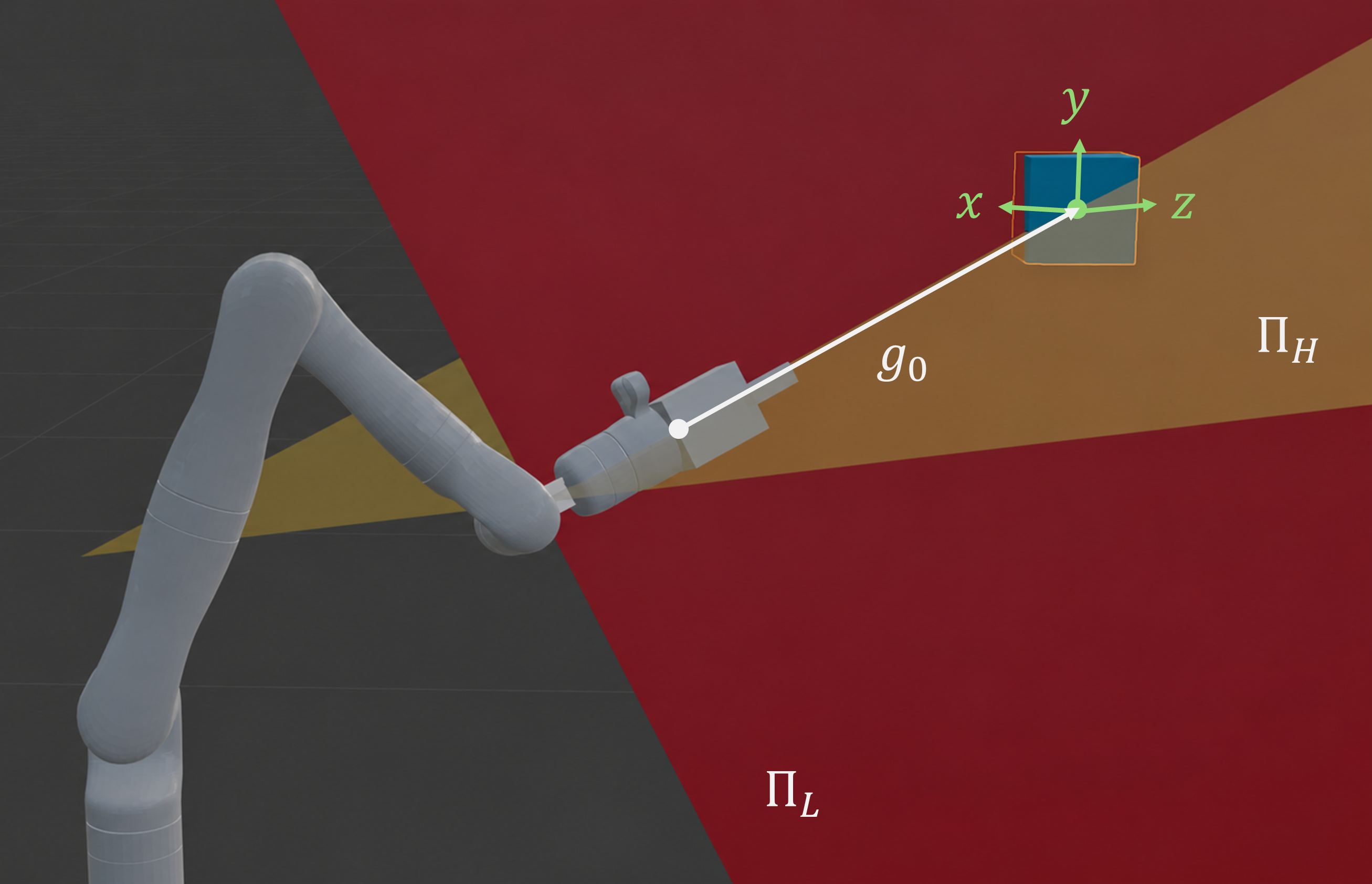}
    \caption{End-effector planes}
    \label{fig:GeometryPlanes}
  \end{subfigure}
  \caption{Geometric parameters of a robotic manipulator in a goal-directed
    task. (a)~The goal vector joins the end-effector frame $(\V R_e, \V p_e)$ 
    to the goal frame $(\V R, \V p)$ attached to the target object (Eq.~\eqref{eq:objectvector}).
    The line of sight $\gamma$ is the angle between the end-effector's local $z$ axis 
    and $\V g$ (Eq.~\eqref{eq:LineOfSight}). (b)~The end
    effector is read as the robot's face and the wrist-mounted camera as its
    eyes. During approach and monitoring, the constraint
    $\hat{\V z}_{e,i} = \hat{\V g}_i$, that is $\gamma_i = 0$, holds at
    every pose $i$ of the task (Eq.~\eqref{eq:ZeroLoS}). (c)~The
    mid-transverse (horizontal, tan) plane $\Pi_H$ and the mid-sagittal
    (longitudinal, red) plane $\Pi_L$ of the end-effector are defined at
    pose~$0$ (Eq.~\eqref{eq:Planes}). Their line of intersection lies along the
    line of sight $\hat{\V z}_{e,0} = \hat{\V g}_0$ and passes through the
    goal.}
  \Description{Three rendered views of the same seven-degree-of-freedom robot
    arm and a cube-shaped target. In the left panel the gripper points to the
    side of the cube: axes are drawn at the cube and at the gripper, an arrow
    labeled g runs from the gripper to the cube, and a marked angle gamma
    separates this arrow from a dashed line extending the gripper's forward
    axis. In the middle panel the arm has turned so that the gripper points
    straight at the cube; the gripper's forward axis and the arrow to the cube
    are collinear and the angle is labeled gamma sub i equals zero. In the
    right panel a tan horizontal plane and a dark red vertical plane pass
    through the gripper and intersect along a line that runs from the gripper
    through the cube.}
  \label{fig:Geometry}
\end{figure*}

As shown in Fig.~\ref{fig:GeometryArbitrary}, the local coordinate system
of the goal is given by $(\V R, \V p) \in SE(3)$, expressed in a fixed world
frame, where $\V R = [\V x, \V y, \V z]$ and $\V p = [p_x, p_y, p_z]^T$. The
end-effector frame is denoted by $\{e\} \equiv (\V R_e, \V p_e)$, where
$\V R_e = [\V x_e, \V y_e, \V z_e]$ and $\V p_e$ is the position of the end
effector with respect to the goal. Configurations of the trajectory are
indexed by $i \in \{0, 1, \dots, n\}$ and occur at times $t_i$. We define the
goal vector $\V g$ joining the end-effector to the goal at configuration $i$,
and its unit direction for $\V g_i \neq \V 0$, as

\begin{equation}
  \V{g}_i \coloneqq -\V p_e(t = t_i), \qquad
  \hat{\V g}_i \coloneqq \frac{\V g_i}{\lVert \V g_i \rVert_2}.
  \label{eq:objectvector}
\end{equation}

Robot gaze with respect to the goal is operationalized by \textit{line of
sight} $\gamma$, the angle between the unit vector along the end-effector's
local $z$ axis, $\hat{\V z}_{e,i}$, and the goal direction at configuration
$i$,

\begin{equation}
  \gamma_i \coloneqq \cos^{-1}\!\left(\hat{\V z}_{e,i} \cdot \hat{\V g}_i\right)
  \in [0, \pi].
  \label{eq:LineOfSight}
\end{equation}

The line of sight follows from how observers read the manipulator's body
(Fig.~\ref{fig:GeometryOperational}). Humans attribute intentions to
non-humanoid bodies from their motion alone,
particularly when they need to predict a nonhuman agent's behavior
\cite{Epley2007}, and restrained humanlike cues make machine behavior easier
to interpret \cite{Duffy2003}. On a manipulator, the distal link that orients
toward and contacts objects is the natural site of this attribution.
Terzio\u{g}lu et al. designed gaze and breathing cues for a collaborative arm
and found that such social cues shaped how collaborators perceived the robot
\cite{Terzioglu2020}. We therefore treat the end-effector as the robot's face and
the wrist-mounted camera, whose optical axis is taken as parallel to
$\hat{\V z}_e$, as its eyes, so that $\gamma_i$ is the robot's gaze error.

Holding $\gamma_i = 0$ serves perception, communication and geometry. It keeps
the target at the image center of the eye-in-hand camera, since loss of the
target from view is a principal failure mode of visual servoing
\cite{Chaumette2006}. It  also lets observers read the robot's attention from where
its face points \cite{Admoni2017}, mirroring human eye-hand
coordination, in which gaze reaches contact landmarks ahead of the hand
\cite{Johansson2001}. As a result, changes in motion are then attributed to the robot's
assessment of its goal. It also fixes two rotational degrees-of-freedom (DoF) of the
end-effector as a function of its position, leaving only roll about
$\hat{\V z}_e$, so that every change in
bearing appears as a turn of the robot's head.

For expressive motion, the robot's direct perception of the target object is
therefore initiated by first bringing it to pose~$0$, at which
$\gamma_0 = 0$. Approach and monitoring begin from this pose, and the line of
sight is kept at zero throughout the task,

\begin{equation}
  \gamma_i = 0 \iff \hat{\V z}_{e,i} = \hat{\V g}_i
  \qquad \forall i \in \{0, 1, \dots, n\}.
  \label{eq:ZeroLoS}
\end{equation}

Pose~$0$ also fixes the reference planes for gaze. With a plane given by a
point $\V a$ and a normal $\V n$, $\Pi \equiv (\V a, \V n)$, the mid-transverse
(horizontal) and mid-sagittal (longitudinal) planes of the end-effector
(Fig.~\ref{fig:GeometryPlanes}) are

\begin{equation}
  \Pi_H \coloneqq (\V 0, \V y_{e,0}), \qquad
  \Pi_L \coloneqq (\V 0, \V x_{e,0}).
  \label{eq:Planes}
\end{equation}

Because $\gamma_0 = 0$, $\V g_0$ is orthogonal to both normals, so the planes,
specified through the goal origin, also pass through the end-effector at
pose~$0$ and intersect along $\hat{\V g}_0$. They remain fixed during the
task.

Let $\V q(t) \in \mathbb{R}^m$ be the configuration of an $m$-joint
manipulator and $\V J(\V q) \in \mathbb{R}^{3 \times m}$ the translational
Jacobian of the end-effector, so that the end-effector velocity is
$\V v = \V J(\V q)\dot{\V q} = \dot{\V p}_e = -\dot{\V g}$. Following the
operational-space formulation \cite{Khatib1987}, and with $\hat{\V g}$ the
unit goal direction of Eq.~\eqref{eq:objectvector}, we resolve $\V v$
into the component along the instantaneous line to the goal and the component
orthogonal to it,

\begin{align}
  \V v^{\parallel} &\coloneqq (\V v \cdot \hat{\V g}) \, \hat{\V g},
    \label{eq:RadialComponent} \\
  \V v^{\perp} &\coloneqq \V v - \V v^{\parallel},
    \label{eq:TangentialComponent}
\end{align}

so that $\V v = \V v^{\parallel} + \V v^{\perp}$ with
$\V v^{\parallel} \cdot \V v^{\perp} = 0$ by construction. The decomposition
is pointwise orthogonal and therefore carries no cross term,

\begin{equation}
  \lVert \V v \rVert_2^2 =
    \lVert \V v^{\parallel} \rVert_2^2 + \lVert \V v^{\perp} \rVert_2^2 .
  \label{eq:Pythagoras}
\end{equation}

The two components have disjoint task meaning. The radial component changes
the range to the goal and nothing else. Its signed magnitude is the
\textit{task-progress rate}

\begin{equation}
  \rho \coloneqq \V v \cdot \hat{\V g}
    = -\frac{\mathrm{d}}{\mathrm{d}t} \lVert \V g \rVert_2 ,
  \label{eq:ProgressRate}
\end{equation}

with $\rho > 0$ closing range, $\rho < 0$ opening it and $\rho = 0$ holding
it. The tangential component changes the bearing to the goal and nothing
else. Since $\dot{\hat{\V g}} = -\V v^{\perp} / \lVert \V g \rVert_2$, the
angular velocity of the goal vector, which we call the
\textit{target-bearing rate}, is

\begin{equation}
  \V \omega \coloneqq \hat{\V g} \times \dot{\hat{\V g}}, \qquad
  \lVert \V \omega \rVert_2 =
    \frac{\lVert \V v^{\perp} \rVert_2}{\lVert \V g \rVert_2} ,
  \label{eq:BearingRate}
\end{equation}

which vanishes exactly when the motion is purely radial.
The pair $(\rho, \V \omega)$ is thus an exact reparameterization of $\V v$
given $\V g$ -- task-progress rate and target-bearing rate together recover the
velocity, and neither is recoverable from the other. For a manipulator whose
Jacobian has full row rank away from singularities, any admissible
$(\rho, \V \omega)$ is realizable by some $\dot{\V q}$, so the two are
independently commandable. Relative to a fixed goal, end-effector translation
therefore separates into a radial channel, $\rho$, with one DoF,
and a tangential channel, $\V \omega$, with two. Any modulation of
translational motion quality during a goal-directed task acts through the
task-progress rate, the target-bearing rate, or both.

This goal-relative grouping, rather than the Cartesian axes, carries
task meaning, and it matches the dimensions of the $C$-$V$ model. The radial channel carries the robot's motor \textit{commitment} to the goal -- it is the only component that advances toward or
withdraws from the target. The tangential channel carries the robot's
allocation of motion to \textit{vigilance} -- it changes the vantage point
from which the target is observed without changing the range. Null-space
self-motion of kinematically redundant manipulators, which leaves the end-effector position unchanged, is treated
separately in Section~\ref{subsec:ExpressiveDescriptors}.

\subsection{Primitives and Descriptors for Uncertainty Expression}
\label{subsec:ExpressiveDescriptors}

\subsubsection{Motion Primitives}

With Commitment and Vigilance realized as the task-progress
rate and the target-bearing rate, the
same decomposition enumerates the motion primitives which constitute the canonical
behavioral states of Section~\ref{subsec:CVModel}. A primitive
is a temporally bounded segment of the end-effector trajectory of a robotic
manipulator in a goal-directed task, illustrated in
Fig.~\ref{fig:Geometry}, over which the pair $(\rho, \V \omega)$ holds a
fixed qualitative form. Since $\rho$ is a signed scalar and $\V \omega$ either
vanishes or does not, the task-space cases are exhausted by
$\mathrm{sgn}(\rho) \in \{+, 0, -\}$ crossed with
$\lVert \V \omega \rVert_2 \in \{0, {>}0\}$, and four of the six resulting
cases are elemental:

\begin{itemize}
    \item \textbf{Approach} ($\rho > 0$, $\V \omega = \V 0$) is the forward
    translation of the end-effector along its line of sight to the target,
    $\hat{\V z}_{e,i} = \hat{\V g}_i$, which at pose~$0$ is the line of
    intersection of $\Pi_H$ and $\Pi_L$.
    \item \textbf{Pause} ($\rho = 0$, $\V \omega = \V 0$) is the cessation of
    end-effector translation while the agent conducts environmental
    assessment.
    \item \textbf{Retreat} ($\rho < 0$, $\V \omega = \V 0$) is the backward
    translation along the same line. It is the exact spatial inverse of
    approach, differing only in the sign of $\rho$.
    \item \textbf{Probe} ($\rho = 0$, $\V \omega \neq \V 0$) is the lateral or
    arc-like motion around the target at constant range, on either the
    mid-transverse plane $\Pi_H$ or the mid-sagittal plane $\Pi_L$, with the
    end-effector turning to keep $\gamma_i = 0$.
\end{itemize}

The two remaining cases, $\rho \neq 0$ with $\V \omega \neq \V 0$, are
superpositions of approach or retreat with probe and introduce no new
elemental form. A fifth primitive lies outside this classification. The
manipulator can be kinematically redundant for a positioning task, so
$\V J(\V q)$ has a null space $\mathcal{N}(\V J)$ of dimension
$m - \mathrm{rank}\,\V J$, and any $\dot{\V q} \in \mathcal{N}(\V J)$ produces
internal self-motion with $\V v = \V 0$
\cite{Liegeois1977,Khatib1987}. Such motion is invisible to
$(\rho, \V \omega)$ but visible to an observer, because the arm reconfigures
or the end-effector rotates while its position is held. Under
Eq.~\eqref{eq:ZeroLoS} the admissible self-motion is further restricted
to joint velocities that also leave $\hat{\V z}_e$ unchanged. 

\begin{itemize}
    \item \textbf{Twitch} ($\dot{\V q} \in \mathcal{N}(\V J)$) is the null-space self-motion 
    that keeps $\hat{\V z}_e$ unchanged with $\rho = 0$ and $\V \omega = \V 0$.
\end{itemize}

While at low frequency and large amplitude twitch can appear as head tilts (treating the end-effector as the robot's face), 
at high frequency and low amplitude it can appear as a shiver.
Because it produces no end-effector translation, it cannot be written as a
superposition of the other four primitives. The set \{\textit{Approach, Pause, Retreat, Probe, Twitch}\} is therefore minimal, in that no member is a superposition of the
others, and complete for the motions considered here, in that the four
task-space cases cover every admissible $\V v$ and the fifth covers
null-space motion that preserves the line-of-sight constraint.

Approach and retreat are governed by \textit{Commitment}. Time factor controls the
temporal acceleration profile of $\rho$ and Weight factor controls its magnitude,
with retreat activating when $C$ becomes negative or when approach-avoidance
repetition occurs at $C \approx 0$. In LMA terms, approach realizes the
action-drive component, the robot's resolved motor intention
directed toward the target. Probe is governed by \textit{Vigilance}, 
being the multi-focused scanning mode that drives the
agent to sample from multiple vantage points, and the primitive that
makes $\V \omega$ nonzero. Pause reflects a shift away from the action drive suspending 
goal-directed motion, and can be a manifestation of either Confidence or Vigilance,
depending on low or high readiness to halt. Finally, twitch is a kinematic expression of extreme Vigilance.

Each canonical behavior is a composition of these five primitives.
\textit{Confidence} keeps $\rho \geq 0$
throughout with $\V \omega \approx \V 0$ and no null-space activity, so it is
dominated by approach -- a direct forward trajectory with at most brief pauses
and no scanning, retreat or twitch. \textit{Curiosity} keeps $\rho \geq 0$
while sustaining $\lVert \V \omega \rVert_2 > 0$, superposing probe on
approach and interleaving intervals of $\rho = 0$, $\V \omega = \V 0$. As a result, the robot
approaches while monitoring and converges on the target from multiple angles.
\textit{Hesitance} lets $\rho$ change sign repeatedly, with pause intervals between
advances and withdrawals -- the approach-avoidance conflict in which the agent
tentatively advances, freezes, withdraws partially and may repeat. \textit{Fear} shifts the balance further toward withdrawal -- net
progress is small or negative, withdrawal is abrupt, and the robot holds
still between movements with high-frequency oscillation, a pattern that
parallels the freezing and flight responses of animals and humans under
threat \cite{Roelofs2017}. \textit{Inactivity} has $\V v \equiv \V 0$ and
$\dot{\V q} \equiv \V 0$, the empty composition, and hence not realizable as a distinguishable
expressive stimulus for perceptual uncertainty.

\subsubsection{Expressive Descriptors}

From the five motion primitives, we derive a set of computable descriptors that
parametrically encode the observable kinematics of
uncertainty-modulated manipulator behavior. Each descriptor is a
scalar-valued functional of one of the three quantities the decomposition
exposes, namely the task-progress rate $\rho$, the target-bearing rate
$\V \omega$, or the null-space joint velocity
$\dot{\V q} \in \mathcal{N}(\V J)$. Given the goal position, each is computable
in real time from joint-encoder data via forward kinematics. 

The configurations $i$ are the key frames of the trajectory. A configuration at which all joint
velocities vanish, $\dot{\V q}(t_i) = \V 0$, and are held for a finite
interval is a pause.
The projection of a vector $\V u$ onto $\Pi$ is represented as $\Upsilon_\Pi(\V u)$.
We write $\mathcal{A}$ and $\mathcal{R}$ for the sets of
key-frame intervals $[t_i, t_{i+1}]$ over which the range to the goal
decreases ($\lVert\V{g}_{i+1}\rVert_2 < \lVert\V{g}_{i}\rVert_2$) and
increases ($\lVert\V{g}_{i+1}\rVert_2 > \lVert\V{g}_{i}\rVert_2$),
respectively. The angle of joint $j$ at configuration $i$ is denoted $\chi_{j,i}$.
The expressive descriptors are defined as follows.

\begin{itemize}

\item \textbf{Approach Acceleration} ($\alpha_{\mathrm{approach}} \in \mathbb{R}_{\geq 0}$)
  [$\mathrm{m/s^2}$] is the peak magnitude of the end-effector's translational
  acceleration across all approach episodes $[t_i, t_{i+1}] \in \mathcal{A}$.

\begin{equation}
  \alpha_{\mathrm{approach}} = \max_{[t_i, t_{i+1}] \in \mathcal{A}} \;
    \max_{t \in [t_i, t_{i+1}]} \lVert \ddot{\V p}_e(t) \rVert_2
  \label{eq:ApproachAcceleration}
\end{equation}

\item \textbf{Pause Count} ($n_{\mathrm{pause}} \in \mathbb{N}_0$) [-] is the number of
  interior configurations at which all joint velocities remain zero for at least a duration $\epsilon > 0$,

\begin{equation}
  n_{\mathrm{pause}} = \sum_{i=1}^{n-1} \mathbf{1}[\dot{\V q}(t) = \V 0 \;\; \forall t \in [t_i, t_i + \epsilon]]
  \label{eq:PauseCount}
\end{equation}

\item \textbf{Pause Duration} ($\tau_{\mathrm{pause}} \in \mathbb{R}_{\geq 0}$) [$\mathrm{s}$] is the
  temporal duration of a pause. Within a given trajectory all pauses have
  equal duration.

\item \textbf{Retreat Count} ($n_{\mathrm{retreat}} \in \mathbb{N}_0$) [-] is the number
  of distinct episodes during which the end-effector translates away from the
  goal from configuration $i$ to $i+1$, with the end-effector at rest at both
  configurations.

\begin{equation}
  n_{\mathrm{retreat}} = \sum_{i=0}^{n-1} \mathbf{1}\!\left[
    [t_i, t_{i+1}] \in \mathcal{R}, \;
    \dot{\V q}(t_i) = \dot{\V q}(t_{i+1}) = \V 0 \right]
  \label{eq:RetreatCount}
\end{equation}

\item \textbf{Retreat Distance} ($d_{\mathrm{retreat}} \in \mathbb{R}_{\geq 0}$) [$\mathrm{m}$] is the
  Euclidean distance covered by the end-effector during a retreat episode
  $[t_i, t_{i+1}] \in \mathcal{R}$. Within a given trajectory all retreat
  episodes cover equal distance.

\begin{equation}
  d_{\mathrm{retreat}} = d_{\mathrm{retreat}_i} = \lVert\V{g}_{i+1} - \V{g}_{i}\rVert_2
  \label{eq:RetreatDistance}
\end{equation}

\item \textbf{Retreat Acceleration} ($\alpha_{\mathrm{retreat}} \in \mathbb{R}_{\geq 0}$)
  [$\mathrm{m/s^2}$] is the peak magnitude of the end-effector's translational
  acceleration across all retreat episodes $[t_i, t_{i+1}] \in \mathcal{R}$.

\begin{equation}
  \alpha_{\mathrm{retreat}} = \max_{[t_i, t_{i+1}] \in \mathcal{R}} \;
    \max_{t \in [t_i, t_{i+1}]} \lVert \ddot{\V p}_e(t) \rVert_2
  \label{eq:RetreatAcceleration}
\end{equation}

\item \textbf{Horizontal Gaze} ($\theta_H \in \mathbb{R}_{\geq 0}$) [$^{\circ}$] is
  the amplitude of the angle in the horizontal plane traversed by the end
  effector during each gaze episode. It is computed as the angle between the
  goal vectors at configurations $0$ and $i$ projected onto $\Pi_H$, where
  configuration $i$ is a pause at the end of the gaze episode. Within a given
  trajectory all horizontal gaze episodes cover equal angle.

\begin{equation}
  \theta_{H} = \theta_{H_i} = \cos^{-1} \left(
    \frac{\Upsilon_{\Pi_H}(\V{g}_{0}) \cdot \Upsilon_{\Pi_H}(\V{g}_{i})}
         {\lVert\Upsilon_{\Pi_H}(\V{g}_{0})\rVert_2 \, \lVert\Upsilon_{\Pi_H}(\V{g}_{i})\rVert_2}
    \right)
  \label{eq:HorizontalGaze}
\end{equation}

\item \textbf{Vertical Gaze} ($\theta_V \in \mathbb{R}_{\geq 0}$) [$^{\circ}$] is the
  amplitude of the angle in the longitudinal plane traversed by the end
  effector during each gaze episode, computed as above but projected onto
  $\Pi_L$. Within a given trajectory all vertical gaze episodes cover equal
  angle.

\begin{equation}
  \theta_{V} = \theta_{V_i} = \cos^{-1} \left(
    \frac{\Upsilon_{\Pi_L}(\V{g}_{0}) \cdot \Upsilon_{\Pi_L}(\V{g}_{i})}
         {\lVert\Upsilon_{\Pi_L}(\V{g}_{0})\rVert_2 \, \lVert\Upsilon_{\Pi_L}(\V{g}_{i})\rVert_2}
    \right)
  \label{eq:VerticalGaze}
\end{equation}

\item \textbf{Tilt Angle} ($\chi_{\mathrm{tilt}} \in \mathbb{R}_{\geq 0}$) [$^{\circ}$] is the
  amplitude of the roll angle of the end-effector about its forward-motion
  axis (the local $z$ axis) during each tilt episode between configurations
  $i$ and $i+1$, analogous to a head tilt. Here $\chi_{e,i}$ denotes the roll
  angle of the end-effector about $\hat{\V z}_e$ at configuration $i$.

\begin{equation}
  \chi_{\mathrm{tilt}} = \chi_{\mathrm{tilt}_i} = |\chi_{e,i+1} - \chi_{e,i}|
  \label{eq:TiltAngle}
\end{equation}

\item \textbf{Tilt Velocity} ($\dot{\chi}_{\mathrm{tilt}} \in \mathbb{R}_{\geq 0}$)
  [$^{\circ}/\mathrm{s}$] is the average angular velocity during each tilt episode.

\begin{equation}
  \dot{\chi}_{\mathrm{tilt}} = \dot{\chi}_{\mathrm{tilt}_i} =
    \left| \frac{\chi_{\mathrm{tilt}_i}}{t_{i+1} - t_i} \right|
  \label{eq:TiltVelocity}
\end{equation}

\item \textbf{Shiver Amplitude} ($\chi_{\mathrm{shiver}_{j}} \in \mathbb{R}_{\geq 0}$)
  [$^{\circ}$] is the peak-to-peak amplitude of high-frequency oscillatory
  motion of joint $j$ about the forward-motion axis, where $j$ is a redundant
  joint not active in the current end-effector movement and $i$, $i+1$ are the
  two peak-amplitude configurations.

\begin{equation}
  \chi_{\mathrm{shiver}_{j}} = |\chi_{j,i+1} - \chi_{j,i}|
  \label{eq:ShiverAmplitude}
\end{equation}

\end{itemize}

This set of computable expressive descriptors provides a systematic method to generate
robot trajectories with motion primitives that can parametrically express 
uncertainty-related behavioral states along their constituent channels.
Approach and retreat acceleration contain the
peak magnitudes of $\dot{\rho}$ over episodes of positive and negative $\rho$
respectively, within which the motion is radial by construction. Retreat
count is the number of maximal intervals of $\rho < 0$ delimited by pauses,
and retreat distance carries $\int |\rho| \, \mathrm{d}t$ over one such interval.
Pause count and duration measure the intervals on which $\rho = 0$ and
$\V \omega = \V 0$ jointly. Horizontal and vertical gaze measure the rotation
of $\hat{\V g}$ accumulated about the normals of $\Pi_H$ and $\Pi_L$, i.e.,
the corresponding components of $\V \omega$. Tilt angle, tilt
velocity and shiver amplitude are functionals of
$\dot{\V q} \in \mathcal{N}(\V J)$. Every component of the kinematic decomposition
therefore has at least one descriptor, and every uncertainty-expressive descriptor measures a
behavioral quantity that the decomposition exposes.


\section{Experiments}
\label{sec:Experiments}

\begin{figure}[tbp]
  \centering
  \includegraphics[width=0.75\columnwidth]{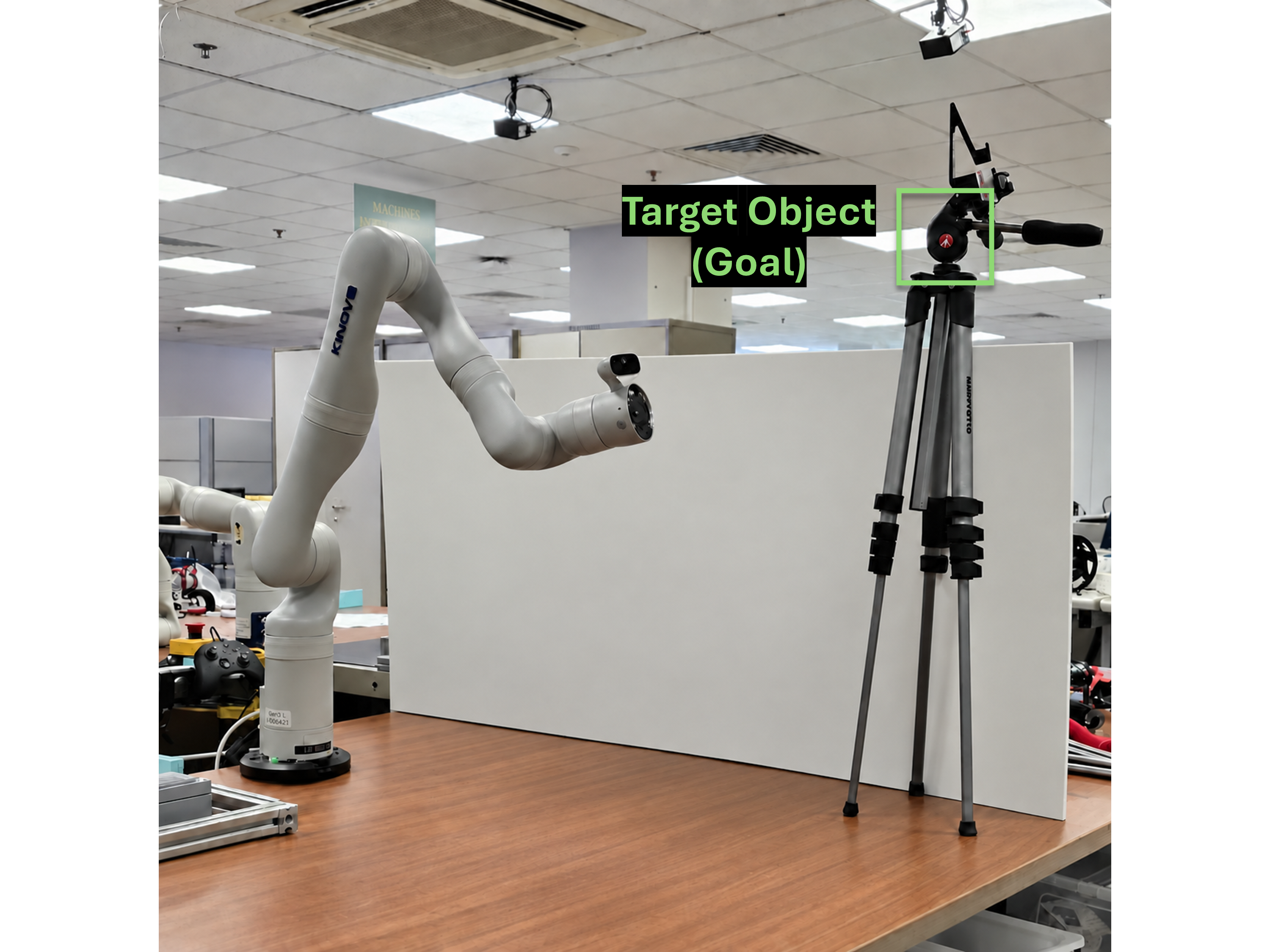}
  \caption{Setup of the 7-DoF Kinova Gen3 Ultralight arm with the target object used in the experiments.}
  \Description{Laboratory setup with a white collaborative robotic arm mounted on the left side of a wooden table. The arm bends upward and extends toward the center, with its circular tool flange facing slightly toward the target object. A tall tripod with a mounting bracket stands on the right, with the target object on top. A single white panel behind the equipment provides a plain backdrop, with laboratory furnishings and ceiling lights visible beyond it.}
  \label{fig:HardwareSetup}
\end{figure}

First, a manipulator trajectory for a goal-directed task, where the end-effector moves to a pre-grasp pose near a target without contact, was populated with combinations of the five motion primitives to elicit four uncertainty-related behaviors. Each trajectory corresponds to one canonical behavioral state that produces observable motion and maps that state to a minimal set of primitives, i.e., Confidence $\mapsto$ \{Approach, Pause\}; Curiosity $\mapsto$ \{Approach, Pause, Probe, Twitch\}; Hesitance $\mapsto$ \{Approach, Pause, Retreat\}; Fear $\mapsto$ \{Approach, Pause, Retreat\}. Next, proof-of-concept demonstrations of uncertainty-expressive motion were conducted on a 7-DoF Kinova Gen3 Ultralight arm (depicted in Fig.~\ref{fig:HardwareSetup}). Following qualitative analysis of the expressivity of the physical motion by three professional movement experts, expressive descriptor sets were tuned to associate with specific behavioral states. Based on this analysis, an IRB-approved (IRB-2025-765), video-based user study was conducted to quantitatively analyze whether descriptor-encoded trajectories communicate uncertainty and whether specific descriptors modulate the perceived intensity of specific uncertainty-related states.

\begin{enumerate}
  \item \textbf{Expression A: Confident.} The confident trajectory reflects complete and
    consistent perception. The motion is direct, with a single pause, no
    retreat, and the highest approach acceleration of the four trajectories
    ($0.1~\mathrm{m/s^2}$, shared with the fearful trajectory).
  \item \textbf{Expression B: Curious.} The curious trajectory represents incomplete
    information about the task and environment. The motion expresses
    information seeking through horizontal gaze shifts of $20^{\circ}$
    separated by two pauses, while the end-effector continues to approach
    the target.
  \item \textbf{Expression C: Hesitant.} The hesitant trajectory corresponds to inconsistent
    or conflicting information. The motion has the most pauses (five), three
    slow retreats of $0.05~\mathrm{m}$, and the lowest approach acceleration
    ($0.05~\mathrm{m/s^2}$).
  \item \textbf{Expression D: Fearful.} The fearful trajectory represents a perceived
    possibility of an unsafe future. The motion combines four pauses with a
    single abrupt retreat (retreat acceleration $2~\mathrm{m/s^2}$, compared
    with $0.6~\mathrm{m/s^2}$ in the hesitant trajectory) and a
    $6^{\circ}$ shiver.
\end{enumerate}

\subsection{Protocol}
\label{subsec:ExperimentalDesign/Protocol}

The study was conducted remotely through Qualtrics, an online survey platform, and participants were 
recruited through posters displayed across university campus and the research group's website.
Participation was voluntary, and responses were collected anonymously without
recording personally identifiable information or IP addresses. Participants
first read an information and consent page describing the purpose of the
research and the tasks. After giving informed consent, they completed a
demographic questionnaire covering age, gender, ethnicity, education level,
and familiarity with and comfort around robots.

In Part A, participants were told that they would watch short video clips of a
robot arm executing expressive behaviors and would judge the robot's
behavioral state. The four base trajectories were identified by participants
only by the letters A to D -- in this study, A is the confident, B the curious,
C the hesitant and D the fearful trajectory. In each Part A trial, the four 
base-trajectory videos were presented together and played in a continuous loop. 
Participants first described the robot's
behavioral state in each video in their own words, before seeing any response
options. They then selected the behavioral state that best described the
robot's behavior from six options -- confident, curious, hesitant, fearful, none
and other. "Other" allowed a different state to be entered, and "None"
indicated that the robot expressed no behavioral state through its motion.
Finally, they rated the intensity of the selected state on a five-point Likert
scale. The stimuli were rendered animations of a new set of base expressive trajectories, whose descriptor values are in Table~\ref{table:PartA}.

\begin{table*}[t]
  \centering
  \caption{Expressive descriptor values of the four base trajectories for Part A of the user study. Descriptors marked N/A, Vertical Gaze, Tilt Angle, and Tilt Velocity are instantiated only in the Part~B variants of the relevant behaviors.}
  \label{table:PartA}
  \Description{A table listing eight expressive descriptors as rows, with
    columns for the symbol, the unit and the value taken by that descriptor in
    each of the four base trajectories.}
  \begin{tabular}{@{}llcrrrr@{}}
    \toprule
    \textbf{Expressive Descriptor} & \textbf{Symbol} & \textbf{Unit} &
      \textbf{Confident} & \textbf{Curious} & \textbf{Hesitant} &
      \textbf{Fearful} \\
    \midrule
    Approach Acceleration & $\alpha_{\mathrm{approach}}$ & $\mathrm{m/s^2}$
      & 0.1 & 0.09 & 0.05 & 0.1 \\
    Pause Count & $n_{\mathrm{pause}}$ & -
      & 1 & 2 & 5 & 4 \\
    Pause Duration & $\tau_{\mathrm{pause}}$ & $\mathrm{s}$
      & 1.67 & 1.25 & 1.67 & 1.25 \\
    Retreat Count & $n_{\mathrm{retreat}}$ & -
      & 0 & 0 & 3 & 1 \\
    Retreat Distance & $d_{\mathrm{retreat}}$ & $\mathrm{m}$
      & N/A & N/A & 0.05 & 0.06 \\
    Retreat Acceleration & $\alpha_{\mathrm{retreat}}$ & $\mathrm{m/s^2}$
      & N/A & N/A & 0.6 & 2 \\
    Horizontal Gaze & $\theta_H$ & $^{\circ}$
      & N/A & 20 & N/A & N/A \\
    Shiver Amplitude & $\chi_{\mathrm{shiver}_{j}}$ & $^{\circ}$
      & N/A & N/A & N/A & 6 \\
    \bottomrule
  \end{tabular}
\end{table*}

In Part B, the relevant descriptors of each base trajectory were modified one
at a time, so that each variant differed from its base trajectory in a single
descriptor. Participants viewed each $\langle$base, variant$\rangle$ pair side
by side and indicated which trajectory, if any, expressed the base behavioral
state with greater intensity. The four options were the trajectory with the
lower value of the manipulated descriptor, the trajectory with the higher
value, "Both appear equally expressive" and "Neither appears expressive".
This gave twenty comparisons -- two for confidence and six each for curiosity,
hesitance and fear. No time limits were imposed. The order of video
presentation was randomized across participants to mitigate order and
learning effects, and all participants viewed the same stimuli. Responses from
participants who did not complete the survey were excluded from analysis. Table~\ref{tab:partB-descriptor-values} contains the descriptors variants for each expressive trajectory used in Part B of the user study.

\begin{table*}[t]
  \centering
  \caption{Descriptor Values for all twenty descriptor comparisons in Part B of the user study. Lower and Higher refer to the lower-valued and higher-valued trajectory, respectively.}
  \label{tab:partB-descriptor-values}
  \Description{A table listing the twenty descriptor comparisons as rows,
    grouped by behavioral state: two for confident and six each for curious,
    hesitant and fearful. Columns give the state, the manipulated expressive
    descriptor, its symbol and unit, and the descriptor value in the
    lower-valued and higher-valued trajectory of each comparison.}
  \begin{tabular}{llllrr}
    \toprule
    \textbf{State} & \textbf{Expressive Descriptor} & \textbf{Symbol} & \textbf{Unit} & \textbf{Lower} &\textbf{ Higher} \\
    \midrule
    Confident & Pause duration & $\tau_{\mathrm{pause}}$ & second $(s)$ & 0.625 & 2.5 \\
    Confident & Approach acceleration & $\alpha_{\mathrm{approach}}$ & $(m/s^2)$ & 0.5 & 1.5 \\
    \addlinespace
    Curious & Pause duration & $\tau_{\mathrm{pause}}$ & second $(s)$ & 0.625 & 2.5 \\
    Curious & Tilt angle & $\chi_{\mathrm{tilt}}$ & degree $(\circ)$ & 0 & 40 \\
    Curious & Horizontal gaze & $\theta_H$ & degree $(\circ)$ & 10 & 27 \\
    Curious & Vertical gaze & $\theta_V$ & degree $(\circ)$ & 15 & 30 \\
    Curious & Approach acceleration & $\alpha_{\mathrm{approach}}$ & $(m/s^2)$ & 0.025 & 0.1 \\
    Curious & Tilt velocity & $\dot{\chi}_{\mathrm{tilt}}$ & degree/s $(\circ/s)$ & 12 & 48 \\
    \addlinespace
    Hesitant & Pause duration & $\tau_{\mathrm{pause}}$ & second $(s)$ & 0.625 & 2.5 \\
    Hesitant & Retreat distance & $d_{\mathrm{retreat}}$ & metre $(m)$ & 0.03 & 0.15 \\
    Hesitant & Tilt angle & $\chi_{\mathrm{tilt}}$ & degree $(\circ)$ & 0 & 32.81 \\
    Hesitant & Retreat count & $n_{\mathrm{retreat}}$ & - & 1 & 5 \\
    Hesitant & Approach acceleration & $\alpha_{\mathrm{approach}}$ & $(m/s^2)$ & 0.025 & 0.075 \\
    Hesitant & Retreat acceleration & $\alpha_{\mathrm{retreat}}$ & $(m/s^2)$ & 0.2 & 0.6 \\
    \addlinespace
    Fearful & Pause duration & $\tau_{\mathrm{pause}}$ & second $(s)$ & 0.625 & 2.5 \\
    Fearful & Shiver amplitude & $\chi_{\mathrm{shiver}_{j}}$ & degree $(\circ)$ & 8 & 12 \\
    Fearful & Pause count & $n_{\mathrm{pause}}$ & - & 2 & 4 \\
    Fearful & Retreat count & $n_{\mathrm{retreat}}$ & - & 0 & 2 \\
    Fearful & Approach acceleration & $\alpha_{\mathrm{approach}}$ & $(m/s^2)$ & 0.025 & 0.1 \\
    Fearful & Retreat acceleration & $\alpha_{\mathrm{retreat}}$ & $(m/s^2)$ & 0.5 & 2.0 \\
    \bottomrule
  \end{tabular}
\end{table*}

Expressive robot motion videos, descriptor variant values for Part B, and study questionnaire are available at the repository \url{https://anonymous.4open.science/r/aou/}.

\subsection{Hypotheses}
\label{subsec:ExperimentalDesign/Hypotheses}

The study tests whether expressive robot motion trajectories can communicate
uncertainty-related states, and whether individual expressive descriptors can
modulate the intensity of those states:

\begin{itemize}
  \item \textbf{H1 (Recognition of behavioral states from expressive robot
    motion).} Each of the four base trajectories will be attributed its
    intended behavioral state (confidence, curiosity, hesitance or fear) at a
    rate significantly above chance.
  \item \textbf{H2 (Effect of expressive descriptors on the intensity of
    expressive behaviors).} Changing a single descriptor will produce a
    significant preference for one of the two trajectories as the more
    intense expression of the base state.
\end{itemize}


\section{Results}
\label{sec:Results}

\begin{figure}[tbp]
  \centering
  \includegraphics[width=\columnwidth]{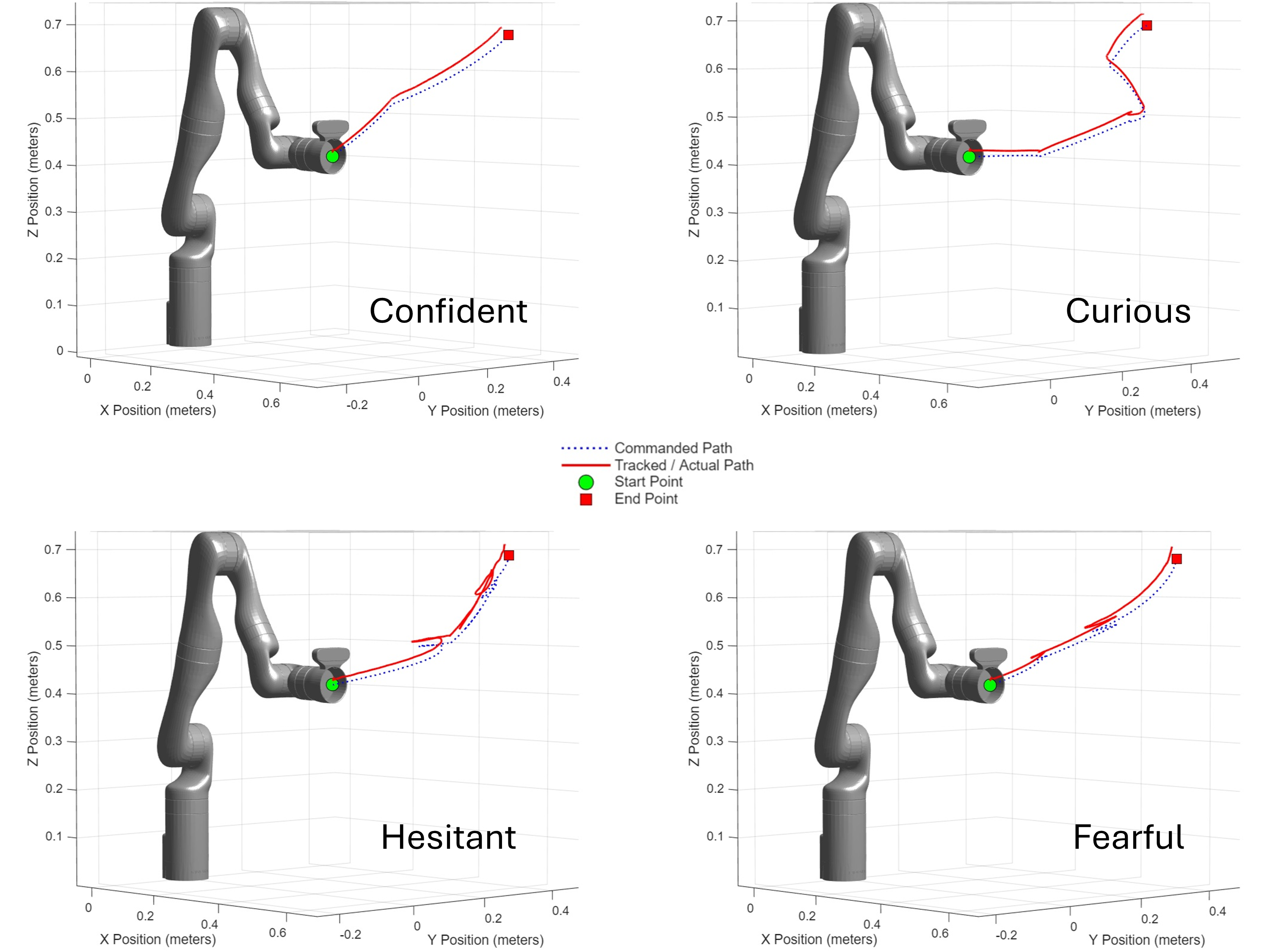}
  \caption{3D position profiles of the end-effector for the four expressive
    trajectories executed on the Kinova Gen3 Ultralight arm for the proof-of-concept experiments. The primitive population distribution for the qualitative analysis is: Confidence $\mapsto$ \{2 Approaches, 1 Pause\}; Curiosity $\mapsto$ \{2 Approaches, 4 Pauses, 2 Probes, 1 Twitch\}; Hesitance $\mapsto$ \{6 Approaches, 5 Pauses, 3 Retreats\}; Fear $\mapsto$ \{4 Approaches, 4 Pauses, 2 Retreats\}.}
  \Description{Four 3D plots compare commanded (blue dotted) and actual (red solid) end-effector trajectories for a seven-degree-of-freedom Kinova Gen3 Ultralight robot: confident (upper left), curious (upper right), hesitant (lower left), and fearful (lower right). Green circles mark starting positions, and red squares mark endpoints. The confident path rises directly, the curious path includes pronounced bends, and the hesitant and fearful paths show localized reversals. Actual paths generally follow commanded paths with visible deviations. All position axes are in meters.}
  \label{fig:HardwareExpressiveTrajectories}
\end{figure}

\begin{figure}[tbp]
  \centering
  \includegraphics[width=\columnwidth]{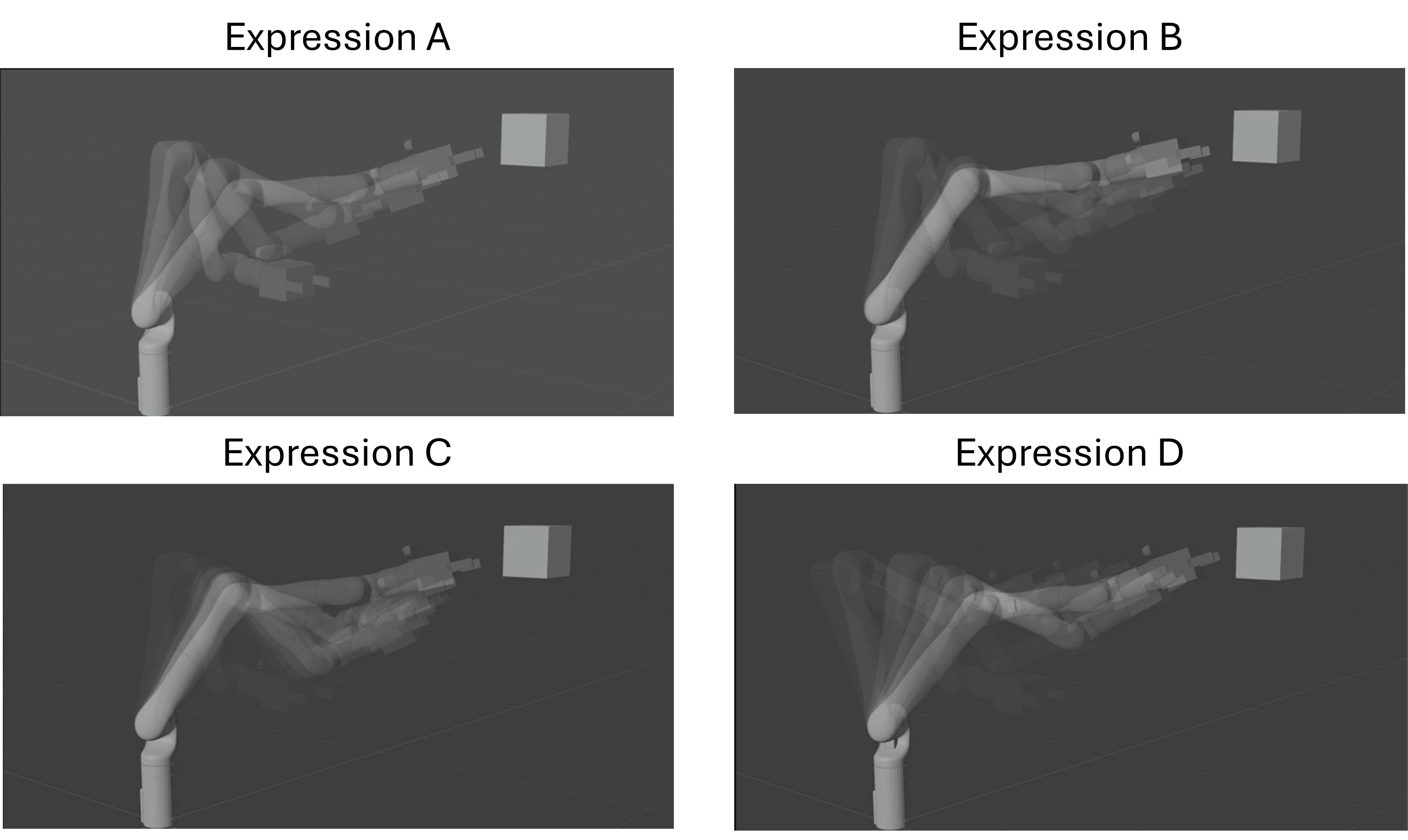}
  \caption{Superimposed manipulator poses sampled at key frames along the four base
    expressive trajectories of the
    goal-directed task in Part A the user study (A: Confident, B: Curious, C: Hesitant, D: Fearful).}
  \Description{Four rendered panels in a two-by-two grid, labeled
    Expression~A to Expression~D. Each panel shows a seven-degree-of-freedom
    manipulator mounted on the left and a gray cube, the target object, on the
    right. Semi-transparent copies of the arm show successive poses as the end
    effector approaches a pre-grasp pose in front of the cube.}
  \label{fig:ExpressiveTrajectories}
\end{figure}

Fig.~\ref{fig:HardwareExpressiveTrajectories} shows the end-effector position profiles and primitive sets of the expressive trajectories used in the proof-of-concept demonstrations. All four trajectories were positively annotated by the movement experts according to their intended states. The qualitative analysis further assessed certain primitive usages as evident, such as twitch (tilt angle and velocity) for curious behavior, and identified other primitives as necessary additions, such as twitch (shiver amplitude) for fearful behavior. As a result, a minimal descriptor set for each intended behavioral state was inductively determined, which provided the input descriptor setting (listed in Table~\ref{table:PartA}) for Part A of the subsequent user study. Fig.~\ref{fig:ExpressiveTrajectories} illustrates the key frames from the four base expressive trajectories used in Part A of the user study.

In the user study, total of 74 survey responses were collected, of which 18 incomplete and 1 with inconsistent demographics were excluded, leaving $N = 55$ valid responses for all analyses. Respondents were 35 male, 19 female, and 1 who preferred not to say. Most were in their twenties (31), with 1 under 20, 16 aged 30–40, 2 aged 40–50, 4 aged 50–60, and 1 over 60. Ethnicity was predominantly Indian (31) and Chinese (14), with 2 Eurasian, 1 Malay, and 7 other. Highest educational attainment was High School (5), Bachelor's (21), Master's (21), and Doctorate or equivalent professional degree (8). Familiarity with robots was mostly moderate (22) or slight (18), with 6 very familiar, 5 extremely familiar, and 4 not familiar. Comfort ratings skewed higher, with 24 moderately comfortable, 15 very, 11 slightly, 3 extremely, and 2 not comfortable.

Chi-squared goodness-of-fit tests assume a uniform distribution across
response categories, and effect sizes are reported as Cohen's $w$
\cite{Cohen1988}. Post-hoc standardized (Pearson) residuals $r$ are
interpreted against a threshold of $\pm 1.96$ \cite{Sharpe2015}, with only positive
residuals being interpreted, since a positive residual identifies a category
selected more often than chance. Because a significant goodness-of-fit test in
Part B can arise from the "Neither" option being avoided rather than from a
preference between the two trajectories, the direction of each Part B effect
was tested with an exact two-sided binomial test on the responses that chose
one of the two trajectories. The twenty binomial $p$-values were corrected
with the Holm-Bonferroni procedure \cite{Holm1979}. Full response counts and
test statistics are given in the supplementary material.

\subsection{Part A: Recognition of Behavioral States}
\label{subsec:ResultsPartA}

\begin{figure}[htbp]
  \centering
  \includegraphics[width=\linewidth]{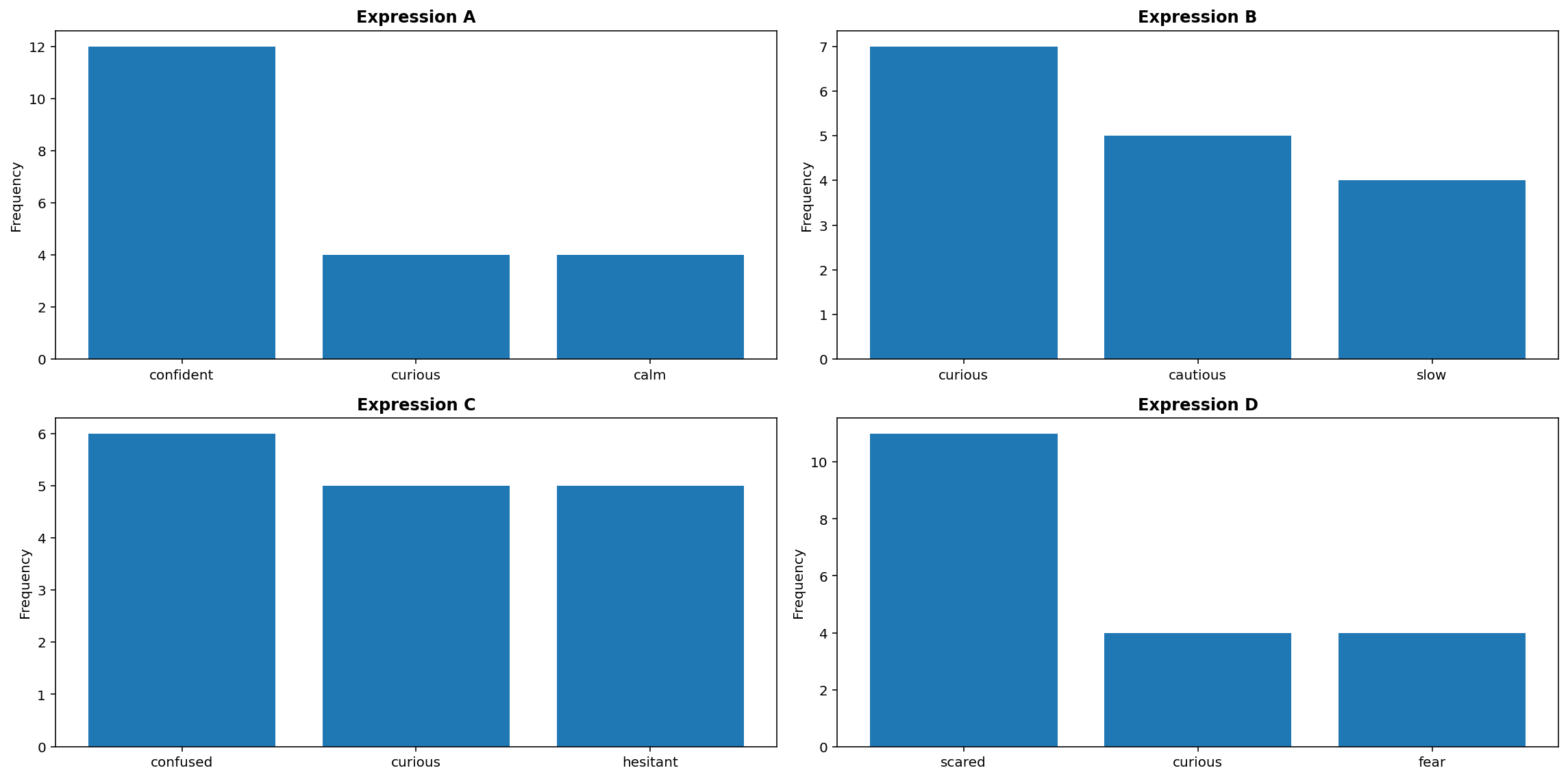}
  \caption{The three most frequent words used by participants to describe each
    base expressive trajectory for the goal-directed task in Part A of the user study (A: Confident, B: Curious, C: Hesitant,
    D: Fearful).}
  \Description{Four small bar charts, one per expression, each showing the
    three most frequently used free-text words and their counts.}
  \label{fig:PartA_BarGraphs}
\end{figure}

\begin{figure}[htbp]
  \centering
  \includegraphics[width=\linewidth]{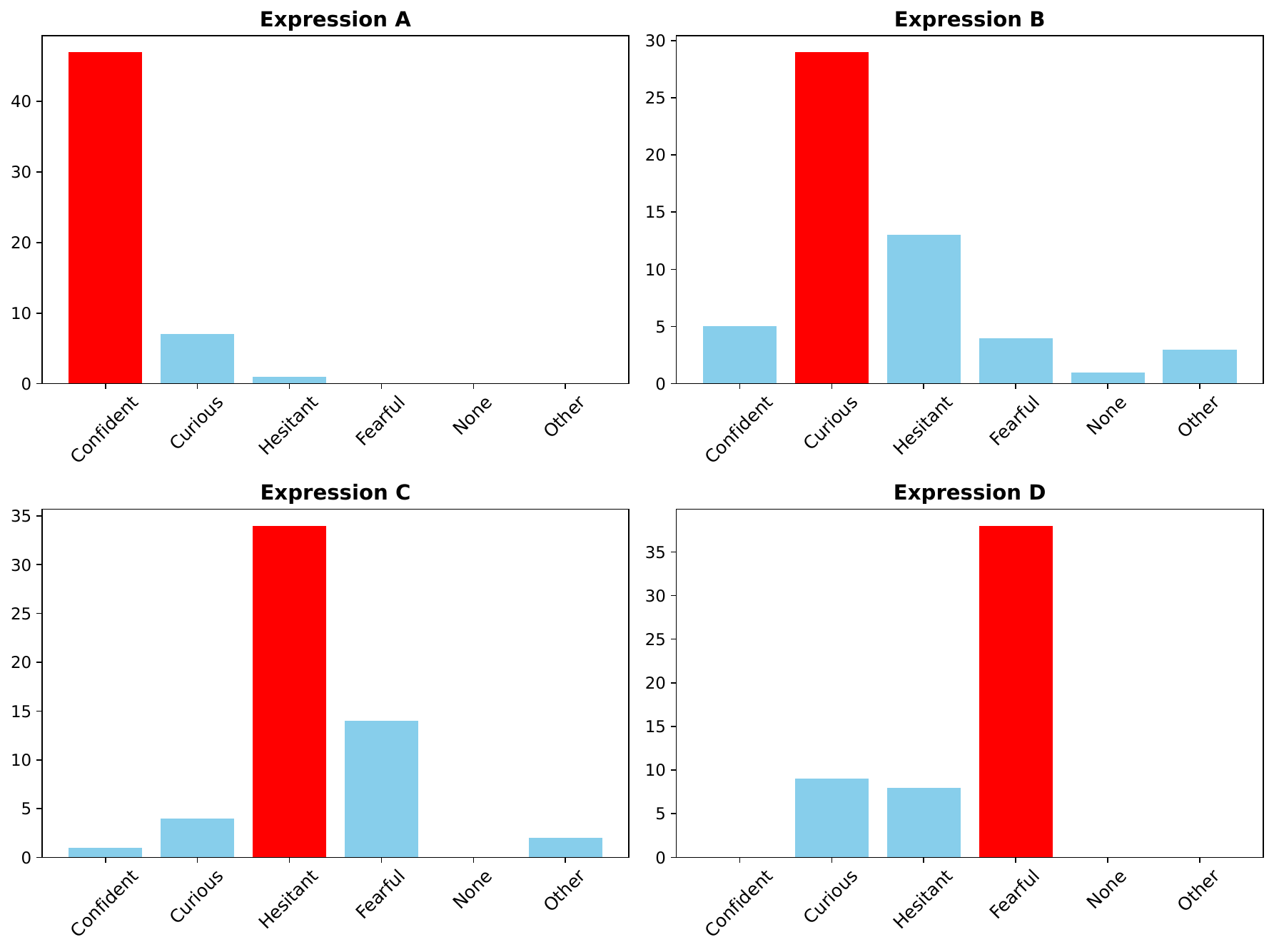}
  \caption{Behavioral states assigned by participants to each base expressive
    trajectory for the goal-directed task in Part A of the user study (A: Confident, B: Curious, C: Hesitant, D: Fearful). The
    intended state is highlighted in each panel.}
  \Description{Four bar charts, one per expression, showing how many of the 55
    participants selected each of the six response categories. In every panel
    the tallest bar corresponds to the intended behavioral state.}
  \label{fig:PartA}
\end{figure}

Fig.~\ref{fig:PartA_BarGraphs} summarizes the free-text descriptions given
before the forced choice, and Fig.~\ref{fig:PartA} shows the forced-choice
responses on which the following tests are based. Chi-squared goodness-of-fit
tests over the six response categories showed significant deviations from
uniformity for all four expressions (Expression~A: $\chi^2(5) = 191.44$;
Expression~B: $\chi^2(5) = 60.75$; Expression~C: $\chi^2(5) = 94.78$;
Expression~D: $\chi^2(5) = 118.35$; all $p < .001$). Effect sizes were large
by Cohen's benchmarks ($w = 1.05$ - $1.87$, against a large-effect threshold of
$w = 0.5$).

Post-hoc standardized residuals identified a single dominant behavioral state
for each expression, in each case the intended one -- \textit{Confident} for
Expression~A ($47$ responses, $r = 12.50$), \textit{Curious} for
Expression~B ($29$ responses, $r = 6.55$), \textit{Hesitant} for Expression~C ($34$ responses,
$r = 8.20$) and \textit{Fearful} for Expression~D ($38$ responses, $r = 9.52$). No other
category within any expression exceeded the threshold. The most frequent
alternative responses were Hesitant for Expression~B ($13$ responses) and Fearful for
Expression~C ($14$ responses).

A chi-squared test of independence over the $4 \times 6$ table indicated that
the distribution of attributed states differed between expressions,
$\chi^2(15) = 261.22$, $p < .001$. Because every participant rated all four
expressions and a third of the expected cell counts are below five, this test
is reported descriptively.

\subsection{Part B: Effect of Expressive Descriptors on Perceived Intensity}
\label{subsec:ResultsPartB}

\begin{figure}[tbp]
  \centering
  \begin{subfigure}[t]{0.48\linewidth}
    \centering
    \includegraphics[width=\linewidth]{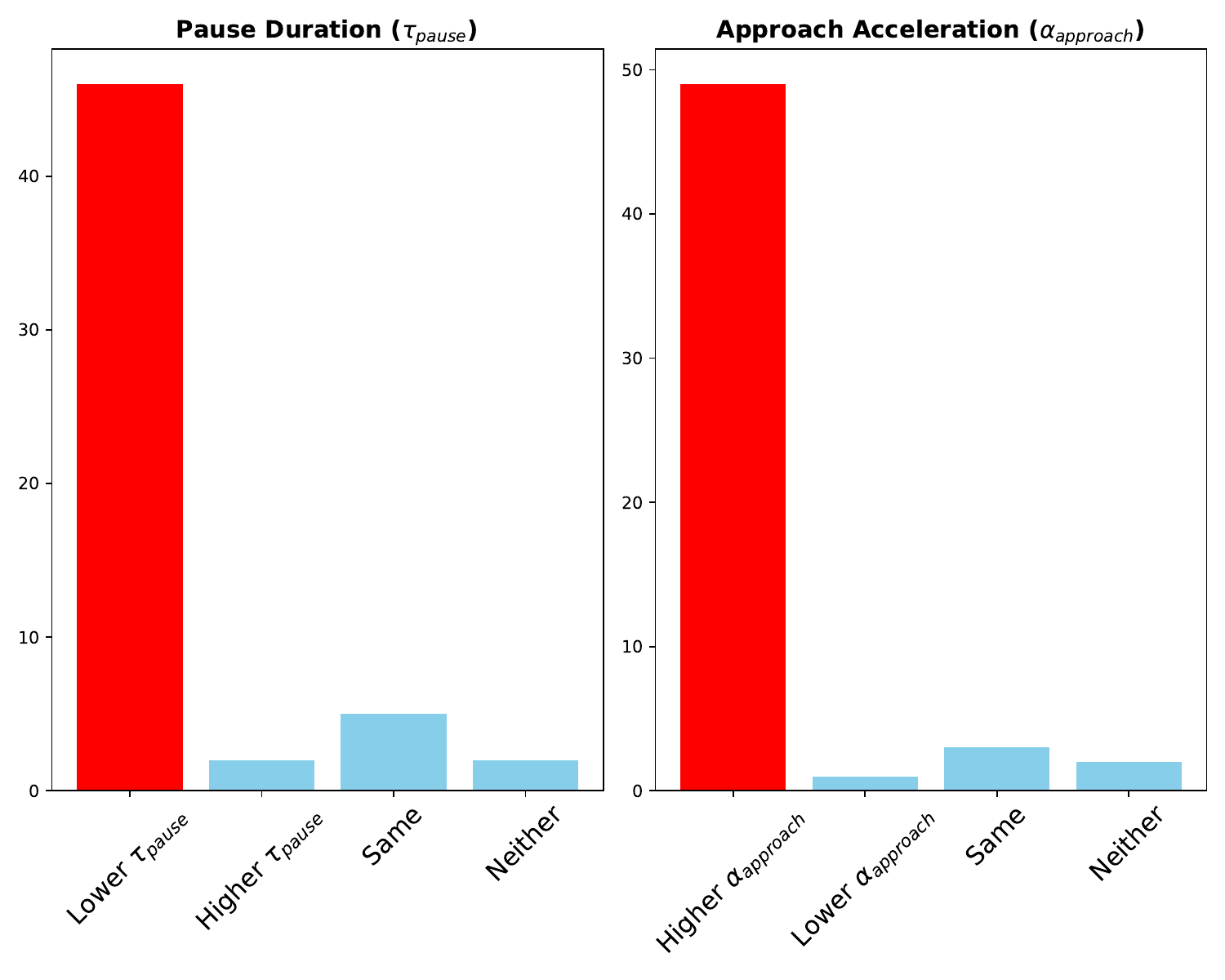}
    \caption{Confident}
    \label{fig:Confidence}
  \end{subfigure}
  \hfill
  \begin{subfigure}[t]{0.48\linewidth}
    \centering
    \includegraphics[width=\linewidth]{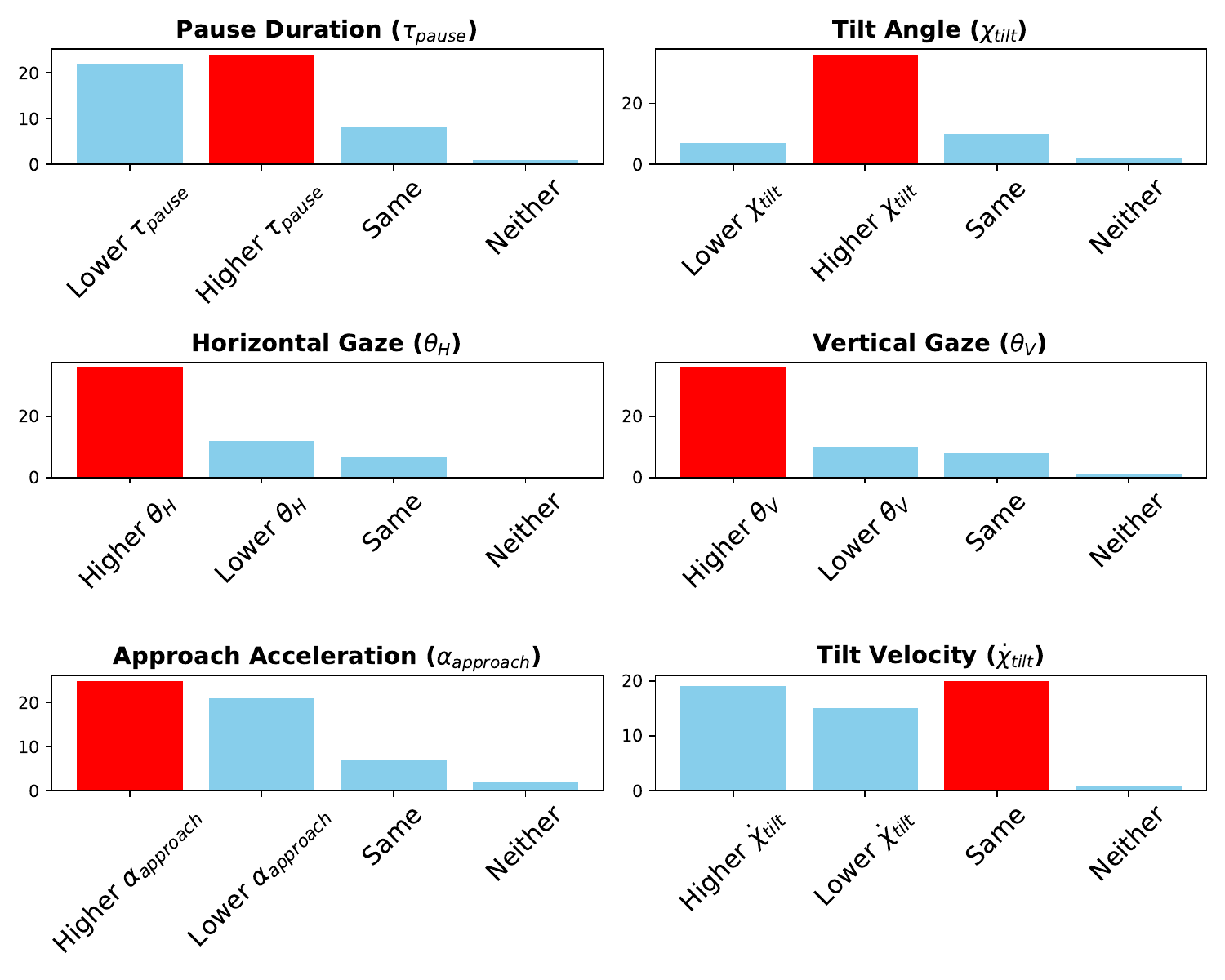}
    \caption{Curious}
    \label{fig:Curiosity}
  \end{subfigure}

  \vspace{0.5em}
  \begin{subfigure}[t]{0.48\linewidth}
    \centering
    \includegraphics[width=\linewidth]{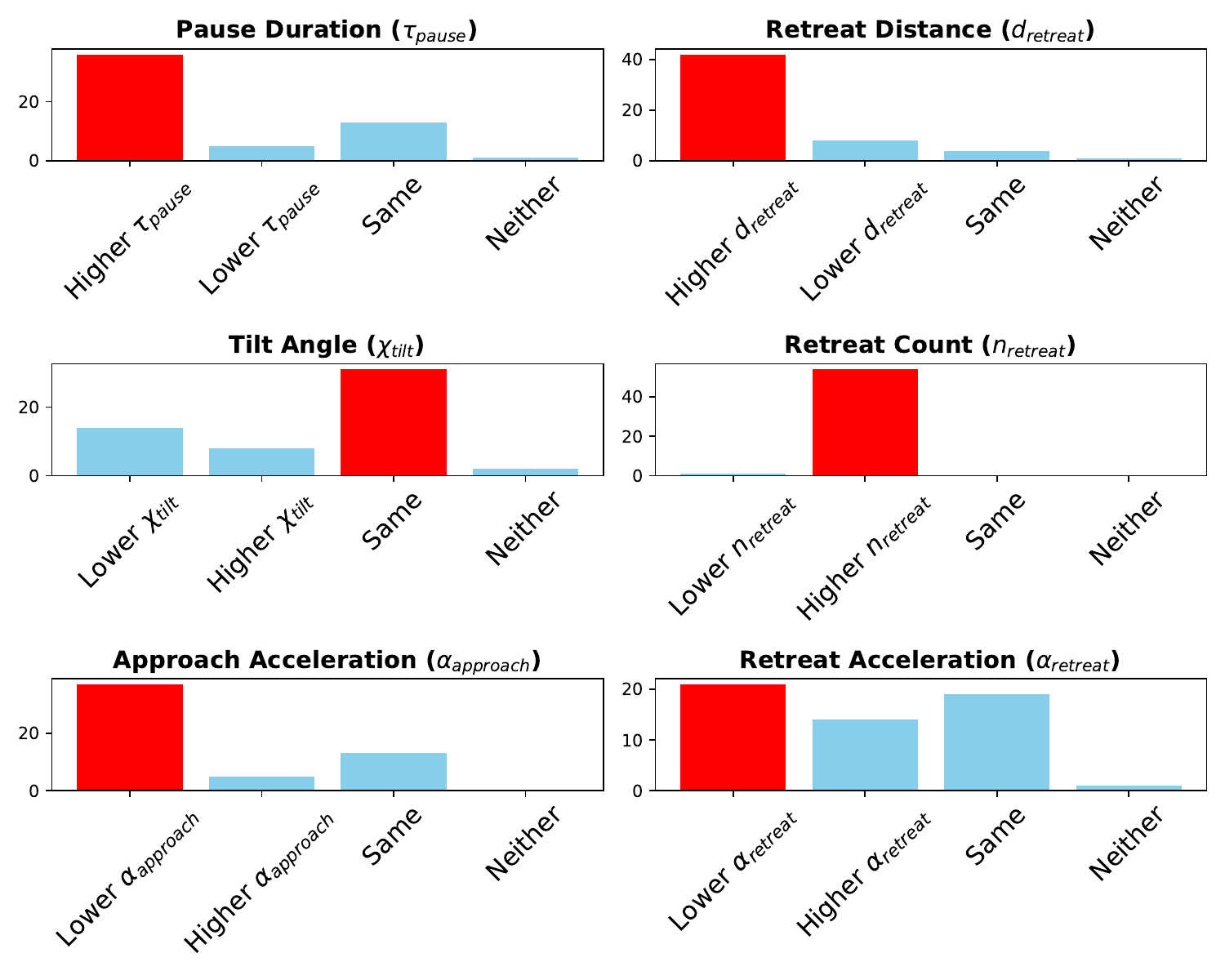}
    \caption{Hesitant}
    \label{fig:Hesitation}
  \end{subfigure}
  \hfill
  \begin{subfigure}[t]{0.48\linewidth}
    \centering
    \includegraphics[width=\linewidth]{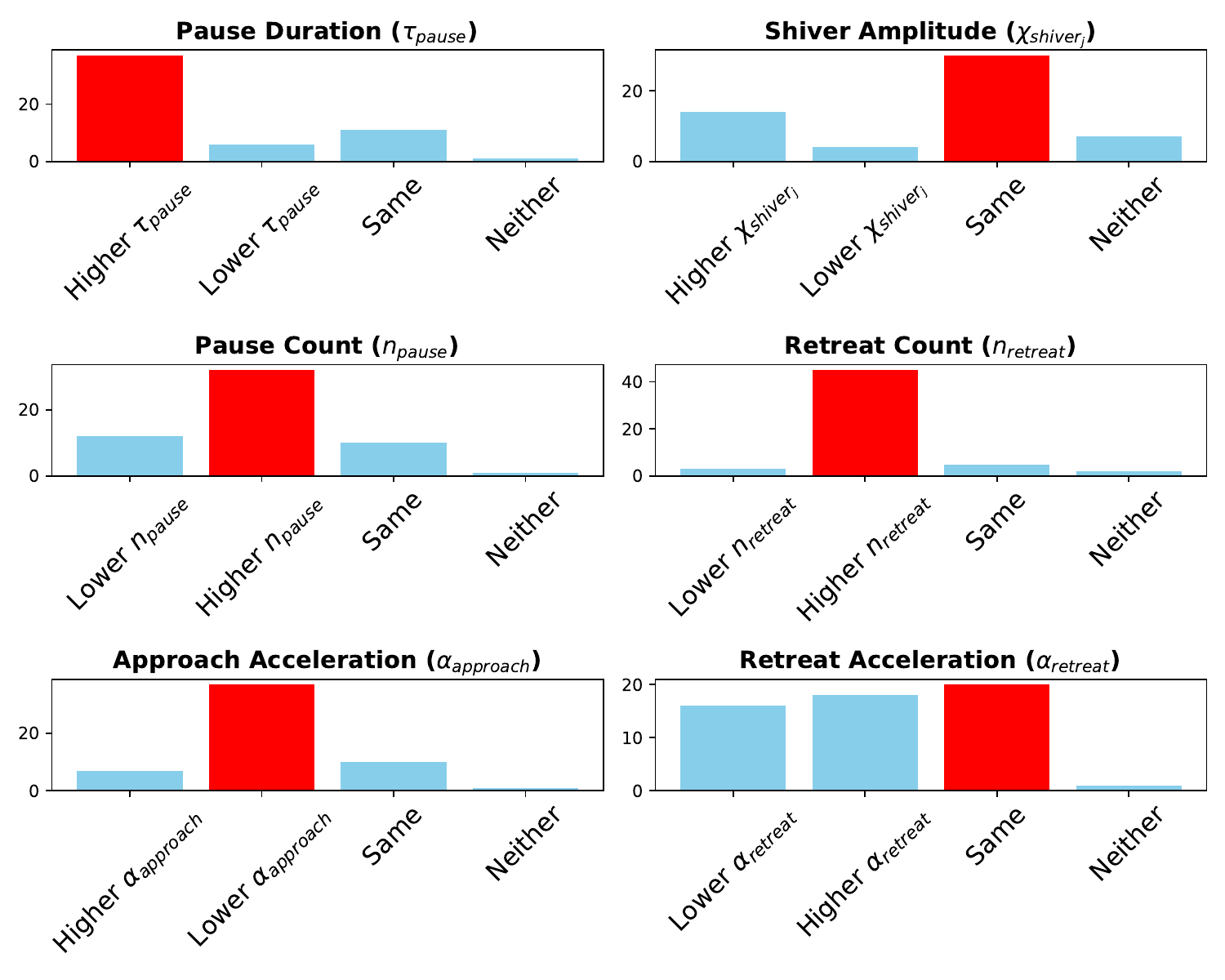}
    \caption{Fearful}
    \label{fig:Fear}
  \end{subfigure}

  \caption{Pairwise intensity judgments for each descriptor manipulation,
    grouped by the four behavioral states ($N = 55$ per chart), in Part B of the user study. In each chart
    the most frequently selected response is highlighted.}
  \Description{Four panels of bar charts. Each small chart corresponds to one
    descriptor and shows how many participants selected the lower-valued
    variant, the higher-valued variant, both equally expressive, or neither.}
  \label{fig:PartB_All}
\end{figure}

Fig.~\ref{fig:PartB_All} shows the response distributions for all twenty
comparisons. All twenty goodness-of-fit tests over the four response options
were significant ($\chi^2(3) \geq 16.35$, all $p < .001$).

\textbf{Confident (Fig.~\ref{fig:Confidence}).} Participants judged the
trajectory with the shorter pause ($46$ vs.\ $2$; $r = 8.70$) and the one with
greater approach acceleration ($49$ vs.\ $1$; $r = 9.51$) as more confident
(both Holm-corrected $p < .001$).

\textbf{Curious (Fig.~\ref{fig:Curiosity}).} Participants judged the
trajectories with greater tilt angle ($36$ vs.\ $7$; $r = 6.00$), greater
horizontal gaze angle ($36$ vs.\ $12$; $r = 6.00$) and greater vertical gaze
angle ($36$ vs.\ $10$; $r = 6.00$) as more curious (Holm-corrected
$p < .001$, $p = .006$ and $p = .002$). Approach acceleration and pause
duration did not produce a preference: responses split between the two
trajectories ($25$ vs.\ $21$ and $24$ vs.\ $22$; both Holm-corrected
$p = 1.00$), and in both cases the two trajectories' residuals were close to
or above the threshold. For tilt velocity no response option exceeded the
threshold.

\textbf{Hesitant (Fig.~\ref{fig:Hesitation}).} Participants judged the
trajectories with longer pauses ($36$ vs.\ $5$; $r = 6.00$), greater retreat
distance ($42$ vs.\ $8$; $r = 7.62$), more retreats ($54$ vs.\ $1$;
$r = 10.85$) and lower approach acceleration ($37$ vs.\ $5$; $r = 6.27$) as
more hesitant (all Holm-corrected $p < .001$). For tilt angle the modal
response was ``both appear equally hesitant'' ($31$; $r = 4.65$). For retreat
acceleration no response option exceeded the threshold.

\textbf{Fearful (Fig.~\ref{fig:Fear}).} Participants judged the
trajectories with longer pauses ($37$ vs.\ $6$; $r = 6.27$), more pauses
($32$ vs.\ $12$; $r = 4.92$), more retreats ($45$ vs.\ $3$; $r = 8.43$) and
lower approach acceleration ($37$ vs.\ $7$; $r = 6.27$) as more fearful
(Holm-corrected $p < .001$, except pause count, $p = .029$). For shiver
amplitude the modal response was "Both appear equally fearful" ($30$;
$r = 4.38$), and among participants who chose a trajectory, more chose the larger
shiver ($14$ vs.\ $4$), although this preference did not survive correction
($p = .22$). For retreat acceleration no response option exceeded the
threshold.

In total, thirteen of the twenty comparisons produced a significant preference
for one trajectory after correction, and in each case the direction of the
preference was consistent with the behavioral model.

\section{Discussion}
\label{sec:Discussion}

The results show that manipulator motion parameterized by the proposed
descriptors can communicate both the type and, for most descriptors, the
intensity of behavioral states associated with perceptual uncertainty.

\subsection{Recognition of Behavioral States}

In Part A of the user study, all four base expressive trajectories deviated significantly from a uniform response
distribution with large effect sizes, and in each case the intended state was
the single dominant response. These results support \textbf{H1}.

Participants' attributions matched the design intent for each trajectory. The
direct trajectory with a single pause and fast approach was identified as
confident, the trajectory with gaze shifts during approach as curious, the
trajectory with repeated pauses, slow retreats and low approach acceleration
as hesitant, and the trajectory with an abrupt retreat and shiver as fearful.
The free-text responses in Fig.~\ref{fig:PartA_BarGraphs} are consistent with
the forced-choice data: the most frequent words were "confident" for
Expression~A, "curious" for Expression~B and "scared" for Expression~D. For
Expression~C the most frequent word was "confused" rather than "hesitant",
which is consistent with hesitance as an unresolved appraisal in which the
robot has neither committed to nor abandoned the goal.

The four base expressive trajectories share the same start pose, target and task, and differ
only in their descriptor values. The residual confusions follow shared
descriptors -- the curious trajectory was most often confused with hesitance
($13$ responses), and both contain repeated pauses; the hesitant trajectory
was most often confused with fear ($14$ responses), and both contain pauses and retreats.
The study did not test single-descriptor versions of the base trajectories, so
it cannot establish how many descriptors are needed to separate the states.
The confusion pattern does suggest that the states are distinguished by the
descriptors they do not share, such as gaze shifts for curiosity.

\subsection{Modulation of Perceived Intensity}

In Part B of the user study, thirteen of the twenty single-descriptor changes produced a significant
preference for one trajectory, which partially supports \textbf{H2}. The
directions were not specified as separate hypotheses before the study; they
are interpreted here against the behavioral model.

For Confidence, shorter pauses and greater approach acceleration were both
judged more confident. The pattern is consistent with findings that the timing of manipulator motion
changes the robot's apparent confidence \cite{Zhou2017} and that the execution
of task actions can convey a robot's internal uncertainty \cite{Hough2017}.
For Curiosity, a larger tilt angle and larger horizontal and vertical gaze
angles each increased perceived curiosity. This is consistent with curiosity as an information-seeking
state and with active perception as motion allocated to observation
\cite{Bajcsy1988}. Approach acceleration and pause duration did not change
perceived curiosity, which suggests that curiosity is carried mainly by the
Vigilance dimension rather than by the tempo of approach.
Hesitance and Fear both became more intense with longer pauses, more retreats
and lower approach acceleration, and hesitance also with greater retreat
distance. The pattern matches the backward movement
components associated with fear in LMA-based analysis of human emotion
recognition \cite{Melzer2019} and the Retreating Shape quality used for
hesitant manipulator trajectories \cite{Raghu2025}.

\subsection{Ambiguous and Non-Dominant Descriptors}

Seven comparisons produced no significant preference -- pause duration,
approach acceleration and tilt velocity in curiosity; tilt angle and retreat
acceleration in hesitance; and shiver amplitude and retreat acceleration in
fear. Participants either judged both trajectories equally expressive or
divided between them. Two explanations are consistent with the data. These
parameters may be perceptually weak when manipulated in isolation and require
covariation with other descriptors. Alternatively, the difference between the
two values may have been too small to exceed a perceptual threshold. The
design used a single step per descriptor and cannot separate the two
explanations.

The duration over which a descriptor acts may also matter. Approach
acceleration shapes the whole approach and was effective in three of four
states, whereas retreat acceleration and tilt velocity change only brief
segments of the trajectory, and shiver is a low-amplitude motion that may be
difficult to resolve in a video. Rendering, viewing distance and screen size
can all limit the visibility of such short or small motions.


\section{Prospects and Limitations}
\label{sec:ProspectsLimitations}

The $C$-$V$ model is defined without reference to a particular body. Commitment
and Vigilance are derived from Effort factors that describe the quality of a
movement rather than the joints that produce it \cite{Laban1947,Bartenieff1980},
and the canonical states are specified by their position on the uncertainty
continuum. The model therefore serves as a generalized behavioral model of
uncertainty-expressive robot motion, in which the platform determines only how
the two dimensional channels are rendered as motion. Expressive motion has been
generated for wheeled, legged, aerial and manipulator embodiments
\cite{Venture2019}, and a morphology-independent specification allows a
behavioral state to be defined once and realized on each of them.

The kinematic decomposition carries this property into trajectory space.
Eqs.~\eqref{eq:ProgressRate} and~\eqref{eq:BearingRate} require only a
controlled point and its position relative to the goal, not a manipulator
Jacobian. For a wheeled base, $\rho$ and $\V \omega$ reduce to the range and
bearing rates of the goal-relative polar coordinates used in unicycle steering
control \cite{Aicardi1995}, with a single tangential degree of freedom in the
plane. For an aerial robot, both tangential degrees of freedom remain available,
and flight paths have already been represented as sequences of motion primitives
whose trajectories and velocities were modified to communicate intent
\cite{Szafir2014}. Approach, pause, retreat, and probe, together with the
descriptors that parameterize them, are therefore defined on any platform that
translates relative to a goal \cite{JangSher2019}, and the line-of-sight constraint becomes a
heading or camera-pointing constraint. In the present formulation, only two
realizations are specific to the manipulator, namely gaze through the
end-effector orientation and twitch through null-space self-motion, which
additionally requires kinematic redundancy. On a non-redundant platform,
twitch must be rendered through another body channel, such as a head or a
sensor mount.

These aforementioned discussion follows from the structure of the model and have not yet been tested empirically. The descriptor values were set and varied manually for a single rendered manipulator, so the study validates that the descriptors modulate the perceived expression of predefined behavioral states, not that the same $C$-$V$ coordinates produce the same attributions on other platforms.
Although the model links a robot's perceptual uncertainty, the $C$-$V$ state
space, the motion primitives and the observable descriptors, it does not yet
compute the descriptor values required for the robot's current perceptual
condition. The generated motions therefore depend on designer interpretation,
and their generalization across tasks, workspaces and changing uncertainty
levels remains open.

The evaluation was video-based and measures interpretation of predefined
trajectories, not closed-loop interaction in which robot perception, motion
generation and human response influence one another. Robots are more
persuasive and are perceived more positively when physically present than
when shown on a screen \cite{Li2015}, and rendered stimuli also remove the
scale, proximity and safety cues of standing beside a manipulator. In Part~A,
the four expressive trajectories were shown together and the response options
included all four intended states, so participants could have assigned some
labels by elimination, and the uniform chance level of one in six therefore
understates the effective chance level. Lastly, all stimuli held the line of
sight at $\gamma_i = 0$, so the study does not measure how gaze directed away
from the goal would change the attributed state.


\section{Conclusion}
\label{sec:Conclusion}

This paper presented a model for expressing a robot's perceptual uncertainty
through the quality of its motion. Commitment and Vigilance, grounded in
approach-avoidance motivation, appraisal theory and active perception and made
computable through Laban Effort factors, define a state space in which
confidence, curiosity, hesitance, fear and inactivity occupy distinct
positions. Decomposing goal-directed end-effector velocity into a task-progress
rate and a target-bearing rate realizes the two dimensions kinematically and
yields five motion primitives parameterized by eleven descriptors, each
computable from joint-encoder data and the goal position. In a video-based
study with $55$ participants, the intended state was the dominant response for
each of the four trajectories that produce observable motion, and thirteen of
the twenty single-descriptor changes produced a significant preference whose
direction was consistent with the model. The results provide a perceptual basis for
communicating uncertainty through the task motion of robots that lack faces,
speech interfaces or displays, and they reduce the design of such motion to a
small set of measurable parameters organized by a morphology-independent
behavioral model.

Future work will address the limitations identified above. To generate
expressive motion autonomously, estimates of perceptual uncertainty, such as the
epistemic and aleatoric uncertainty of learned vision models \cite{Kendall2017}
or the belief state of a partially observable planner \cite{Kaelbling1998},
may be mapped to $C$-$V$ coordinates and from there to descriptor values. The
expressive trajectories can be represented as Dynamic Movement Primitives
(DMPs), whose support for temporal scaling, spatial
adaptation and goal convergence allows descriptor-conditioned trajectory
generation to be posed as an optimization problem subject to task constraints
and the safety requirements for collaborative applications
\cite{ISO10218-2}. To test the claimed generality, the same $C$-$V$
coordinates may be realized on platforms of different morphology, including
non-redundant arms and mobile bases on which gaze and twitch require other
body channels, and the resulting attributions can be compared across
embodiments. Finally, in-person studies with a physically present robot in a
shared task may evaluate recognition under closed-loop interaction, with graded
descriptor levels to estimate perceptual thresholds for the weaker descriptors.


\bibliographystyle{ACM-Reference-Format}
\bibliography{references}

\end{document}